\PassOptionsToPackage{dvipsnames}{xcolor}

\documentclass[10pt,twocolumn,letterpaper]{article}

\usepackage{iccv}
\usepackage{times}
\usepackage{epsfig}
\usepackage{graphicx}
\usepackage{amsmath}
\usepackage{amssymb}
\usepackage{xcolor}
\usepackage{gensymb}
\usepackage{paralist}

\usepackage{epsfig}
\usepackage{graphicx}
\usepackage{wrapfig}
\usepackage{subfigure}

\usepackage{multirow}
\usepackage{rotating}
\usepackage{booktabs}
\usepackage{tabularx}
\usepackage{colortbl}

\usepackage[pagebackref=true,breaklinks=true,letterpaper=true,colorlinks,bookmarks=false]{hyperref}

\iccvfinalcopy

\def\iccvPaperID{10739}

\ificcvfinal\fi

\newcolumntype{s}{>{\columncolor[gray]{.85}[.5\tabcolsep]}c}

\newcommand{\csection}[1]{
    \vspace{-0.05in}
    \section{#1}
    \vspace{-0.04in}
}

\newcommand{\csubsection}[1]{
    \vspace{-0.02in}
    \subsection{#1}
    \vspace{-0.01in}
}

\newcommand{\xhdr}[1]{\vspace{2.5pt}\noindent\textbf{#1}}
\newcommand{\xhdrflat}[1]{\noindent\textbf{#1}}

\newcommand{\tabref}[1]{Tab.~\ref{#1}\xspace}
\newcommand{\figref}[1]{Fig.~\ref{#1}\xspace}
\newcommand{\secref}[1]{Sec.~\ref{#1}\xspace}

\begin{document}

\title{Waypoint Models for Instruction-guided\\Navigation in Continuous Environments}

\author{
    Jacob Krantz$^{1}$\thanks{Work done during an internship at Facebook AI Research.\newline Correspondence: \href{mailto:krantzja@oregonstate.edu}{krantzja@oregonstate.edu}}
    \quad Aaron Gokaslan$^{2,3}$
    \quad Dhruv Batra$^{2,4}$
    \quad Stefan Lee$^{1}$
    \quad Oleksandr Maksymets$^{2}$ \\
    {\normalsize $^1$Oregon State University} \
    {\normalsize $^2$Facebook AI Research} \
    {\normalsize $^3$Cornell University} \\
    {\normalsize $^4$Georgia Institute of Technology} \\    
    {\tt\small Project Webpage: \href{https://jacobkrantz.github.io/waypoint-vlnce}{https://jacobkrantz.github.io/waypoint-vlnce}}
}

\maketitle
\ificcvfinal\thispagestyle{empty}\fi

\begin{abstract}
Little inquiry has explicitly addressed the role of action spaces in language-guided visual navigation -- either in terms of its effect on navigation success or the efficiency with which a robotic agent could execute the resulting trajectory. Building on the recently released VLN-CE \cite{krantz2020beyond} setting for instruction following in continuous environments, we develop a class of language-conditioned waypoint prediction networks to examine this question. We vary the expressivity of these models to explore a spectrum between low-level actions and continuous waypoint prediction. We measure task performance and estimated execution time on a profiled LoCoBot \cite{locobot} robot. We find more expressive models result in simpler, faster to execute trajectories, but lower-level actions can achieve better navigation metrics by approximating shortest paths better. Further, our models outperform prior work in VLN-CE and set a new state-of-the-art on the public leaderboard -- increasing success rate by 4\% with our best model on this challenging task. 
\end{abstract}

\csection{Introduction}

A long-term goal of instruction-guided visual navigation research is to develop AI for robotic agents that can reliably follow paths described by natural language navigation instructions in new environments. 
Much of the existing work in this domain is robot-agnostic and has focused on highly-abstract simulators where agents navigate by choosing between a small, fixed set of nearby locations that the agent then transitions to deterministically \cite{anderson2018vision,chen2019touchdown, hermann2019learning,ku2020room} -- essentially assuming some underlying robot-specific control system can perform navigation. The Vision-and-Language Navigation (VLN) \cite{anderson2018vision} task  is representative of this class of problem settings. 

In sim-to-real experiments, Anderson et al.~\cite{anderson2020sim} demonstrate that a major performance bottleneck for transferring VLN agents trained in high-level simulators to real robotic systems is producing appropriate sets of nearby locations (or waypoints) to choose from; however, it is infeasible to study waypoint prediction in the discrete, highly-abstract simulator as the agent can only occupy predefined locations. 

\begin{figure}
    \centering
    \includegraphics[width=0.98\columnwidth]{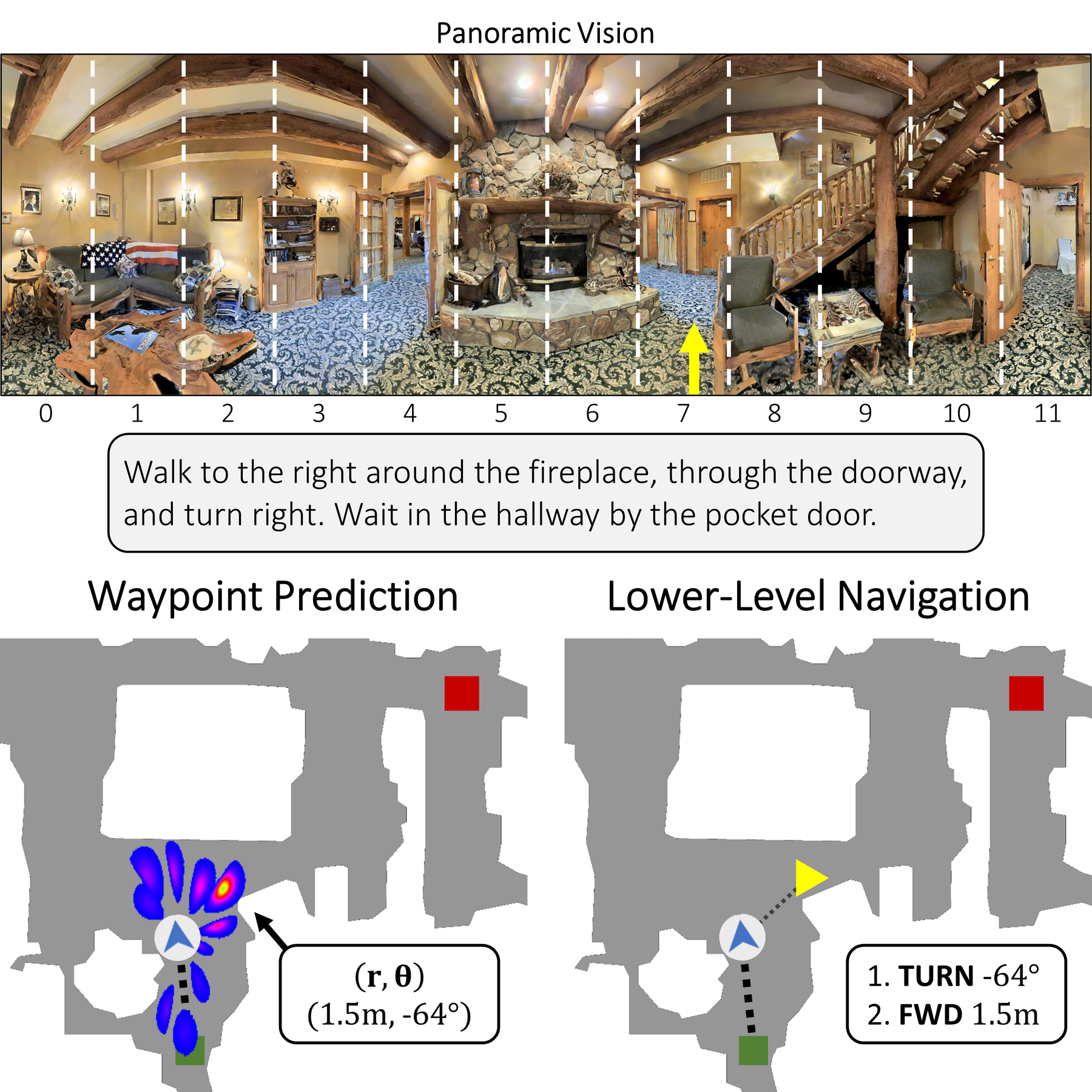}
    \caption{Our approach decomposes the task of following navigation instructions in continuous environments into language-conditioned waypoint prediction and low-level navigation. }
    \label{fig:intro}
\end{figure}

Recently, Krantz et al.~\cite{krantz2020beyond} introduced a variant of VLN instantiated in continuous simulated environments (denoted VLN-CE) such that agents can move to arbitrary positions. In contrast to the highly-abstract action space in VLN, agents in \cite{krantz2020beyond} navigate by executing a sequence of low-level actions such as moving forward 0.25 meters or turning by 15 degree increments. 
This end-to-end, instruction-to-low-level-control design choice has implications both in simulation and for potential sim-to-real transfer to a robotic platform. 
During training, these policies must jointly learn navigation and language grounding over long sequences of actions ($\sim$55 per episode). As a result, \cite{krantz2020beyond} shows that models mirroring successful VLN agents perform substantially worse in VLN-CE. 

On a real robot, the frequent stop, starts, and turns induced by this action space can be slow to execute (requiring frequent changes in velocity and calls to a planner), result in state estimation error, and strain hardware \cite{kadian_sim2realgap_2020, gonzalez2015review}. Further, executing the deep policy network to predict actions at each time step can put extra demand on robot power supplies.

This work explores a spectrum of action spaces between these two extremes -- studying instruction-guided navigators that predict relative waypoints with varied expressivity. At one end, our agents are free to predict relative waypoints as continuous points within some maximum range. On the other, the action space is reduced to taking a fixed step in a direction chosen from a small, finite set of angles -- mimicking \cite{krantz2020beyond} but collapsing consecutive turns. In between, we experiment with mixing discrete and continuous components to parameterize waypoint predictions. 

To do this, we develop an attention-based waypoint prediction network for instruction following. Given a navigation instruction and a panoramic RGBD observation at the current position, our agents predict a distribution over relative waypoints in polar coordinates (consisting of a heading angle $\theta$ and a distance $r$). A low-level continuous navigator is then executed to move in a straight line towards the waypoint -- leaving concerns about obstacle avoidance to the waypoint predictor. We train our agents as model-free control policies using large-scale reinforcement learning~\cite{wijmans2019dd} on the VLN-CE dataset. We evaluate our agents using standard metrics for VLN-CE as well as the estimated execution time for resulting trajectories on a LoCoBot~\cite{locobot} robot.

We find that more expressive waypoint prediction networks result in simpler paths that are faster to execute; however, more constrained action spaces can achieve better performance by more closely approximating shortest paths. Our waypoint models paired with continuous low-level navigators reduce the average estimated time to execute a trajectory by 2.2 times compared to low-level turn/forward models. When paired with discrete low-level navigators to match VLN-CE's action space, our models improve navigation success rate by 1-4\% over prior work on the VLN-CE leaderboard\footnote{\href{http://eval.ai/web/challenges/challenge-page/719}{eval.ai/web/challenges/challenge-page/719}}-- a max relative improvement of 14\%. 

\xhdr{Contributions.} We summarize our contributions as:
\begin{compactitem}[\hspace{3pt}--]
    \item Developing a class of language-conditioned waypoint prediction networks for the VLN-CE task,
    \item Providing empirical analysis of waypoint prediction expressivity's effect on navigation success and estimated time to execute trajectories on a representative robot, 
    \item Demonstrating that our models paired with low-level navigators set a new state-of-the-art on the VLN-CE test leaderboard by an absolute 4\% success rate.
\end{compactitem}
We provide open-source code and pre-trained models at \href{https://github.com/jacobkrantz/VLN-CE}{https://github.com/jacobkrantz/VLN-CE}.

\csection{Related Work}

\noindent\textbf{Instruction-Guided Navigation.}
Many works have examined instruction-guided navigation. 
Task descriptions vary across a number of axes, including
	instruction source (templated \cite{hermann2019learning}, natural language \cite{anderson2018vision}),
	instruction language (monolingual, multilingual \cite{ku2020room}),
	environment setting (indoor, outdoor \cite{chen2019touchdown, hermann2019learning}),
	environment realism (synthetic simulation \cite{misra2018mapping}, realistic simulation \cite{anderson2018vision}, real-world \cite{blukis2020learning, anderson2020sim}),
	navigation affordance (sparse navigation graphs \cite{anderson2018vision}, continuous space \cite{krantz2020beyond, blukis2020learning}),
	and agent (ground-based, quadcopters \cite{blukis2018following}).

One popular task is Vision-and-Language Navigation (VLN)~\cite{anderson2018vision}. VLN has natural language instructions and uses indoor, photo-realistic environments from the Matterport3D dataset \cite{Matterport3D}. A ground-based agent acts on a sparse navigation-graph. In this work, we consider the recently released Vision-and-Language Navigation in Continuous Environments (VLN-CE)~\cite{krantz2020beyond}, a task that lifts VLN to continuous 3D environments. We explore waypoint models that leverage more abstract action spaces in VLN-CE.

\xhdr{Hierarchical Visual Navigation.} Waypoint-based models can be considered a type of hierarchical agent, which has been proposed for many tasks relating to visual navigation. Beyond an intuitive problem decomposition, these are commonly motivated by a desire to carve out self-contained sub-tasks solvable with existing approaches~\cite{bansal2020combining}, circumvent challenges faced by reinforcement learning (RL) algorithms (\eg credit assignment and exploration over long time horizons)~\cite{chaplot2020object, xia2020relmogen}, or to introduce interpretable representations~\cite{eqamodular_corl_2018}. However, these works address embodied navigation tasks that do not condition on language.

More related to our work are those that predict waypoints directly~\cite{bansal2020combining,chaplot2020neural, chen2020waypoints}.  Chaplot \etal~\cite{chaplot2020neural} address the image-goal navigation task with a topological agent that updates a graph with candidate ``ghost nodes", selects a node to navigate to, and performs low-level navigation. Our waypoint-based model differs in that we combine the waypoint prediction and selection steps and condition both with the task goal. Chen \etal~\cite{chen2020waypoints} take a similar approach to ours for audio-visual navigation -- predicting waypoints conditioned on audio goals (e.g.~a teapot whistling) while building a metric map. Our approach predicts waypoints directly from language instructions without a metric map.

Several hierarchical models have been developed for instruction-guided navigation tasks. For an outdoor environment, Misra \etal~\cite{misra2018mapping} decompose the task into goal prediction and action generation. While effective in (nearly) fully-observable environments, this method does not readily transfer to novel environments with partial observability. Likewise, Blukis \etal~\cite{blukis2018mapping} develop a network that predicts and updates a position-visitation distribution en route to the goal. This approach leverages assumptions of an aerial vehicle operating in outdoor environments, namely, nearly full observability compared to an indoor ground-based agent and rare obstacle collisions afforded by aerial free-space. In concurrent work, Irshad \etal~\cite{irshad2021hierarchical} considers discrete action models to be high-level policies and jointly learns a low-level policy for velocity-based continuous control.

Recent sim2real transfer work in VLN has considered adding a software harness that emulates an `online' navigation graph by predicting candidate waypoints~\cite{anderson2020sim}. This mechanism is not conditioned on instructions and just uses local visual / lidar observations. VLN agents trained in topological simulators can then navigate on this graph by invoking a classical navigation stack in the real world. However, these models were found to perform significantly worse than when given a known navigation graph -- suggesting that waypoint prediction remains a bottleneck for VLN transfer. Instead of a two-stage process, we present an alternative -- developing a language-conditioned waypoint prediction network in a continuous simulator.

\xhdr{Training Instruction Followers.}
Many instruction-guided navigation works learn policies via imitation learning~\cite{anderson2018vision, fried2018speaker, blukis2018following, wang2019reinforced, tan2019learning, krantz2020beyond, majumdar2020improving}. Behavior cloning can result in exposure bias. Methods like student forcing and dataset aggregation reduce this but require a queriable expert policy and  discourage exploration~\cite{anderson2018vision, ross2011reduction}. Some works train agents with a combination of imitation learning and reinforcement learning (RL)~\cite{blukis2020learning, ku2020room}. In this work, we learn linguistically-motivated waypoint predictions purely from RL. 

\csection{Task Description}

We consider the episodic task of instruction-guided visual navigation in previously-unseen environments. An agent must navigate a path specified by natural language instructions and stop at a goal location. The agent has egocentric RGBD perception. The environment is continuous, requiring the agent to navigate freely about the 3D space and contend with obstacles and occlusion.

\xhdr{VLN-CE Task.} We set our work in the context of the Vision-and-Language Navigation in Continuous Environments (VLN-CE) task \cite{krantz2020beyond}. VLN-CE is based on the Room-to-Room dataset used in the original VLN task \cite{anderson2018vision}. VLN has agents navigate on a pre-defined graph of viewpoints with scenes from the Matterport3D dataset \cite{Matterport3D}. VLN-CE replaces the viewpoint topology with full Matterport3D scene reconstructions, lifting the VLN task to more realistic navigation in continuous space. We conduct our experiments in VLN-CE because it enables our study of agents that predict arbitrary relative waypoints. We adopt the task settings of VLN-CE with specific extensions detailed below.

\xhdr{Observation Space.} The agent observes RGB and depth images. For both modalities, we extend the 90$\degree$ HFOV of VLN-CE to panoramic 360$\degree$ HFOV. Each panorama is captured as twelve frames angled in 30$\degree$ increments, where each frame has a $90\degree$ HFOV at a resolution of 256 x 256. Panoramic vision is common in related visual navigation tasks like VLN and PointGoal navigation \cite{fried2018speaker, chaplot2020neural} and panoramic sensors could be used in real applications.

\xhdr{Action Space.} Waypoint-based agents can operate independently of the low-level action space used to reach the predicted waypoints. We experiment with two action spaces that operate in discrete time. Specifically, we train and evaluate our agent with continuous-space actions that specify real-valued turn angles and straight-line distances. Such actions can be accomplished by zero-turn-radius robots such as Locobot \cite{gupta2018robot}. Like VLN-CE, we assume perfect actuation to keep results comparable. We also evaluate our agent with the VLN-CE's discrete action space to enable direct comparison with past work (\texttt{forward} 0.25m, \texttt{left} 15 degrees, \texttt{right} 15 degrees, and \texttt{stop}). Actions specifying velocities or accelerations are beyond the scope of this work, but are compatible with waypoint-based agents~\cite{bansal2020combining}.

\csection{Method}

We describe our implementation of a waypoint-based instruction-following agent. A waypoint prediction network (WPN) predicts navigation waypoints or a \texttt{STOP} action directly from pixels and natural language. Waypoints are passed to a lower-level navigator in relative polar coordinates. We employ a simple two-step navigator that turns in the direction of the waypoint then moves forward the predicted distance. This navigator does no direct language processing, separating the task between two agent components.

\csubsection{Waypoint Prediction Network (WPN)}

\begin{figure*}[t]
    \centering
    \includegraphics[
        clip=False,
        trim=0pt 6pt 6pt 7pt,
        width=\textwidth
    ]{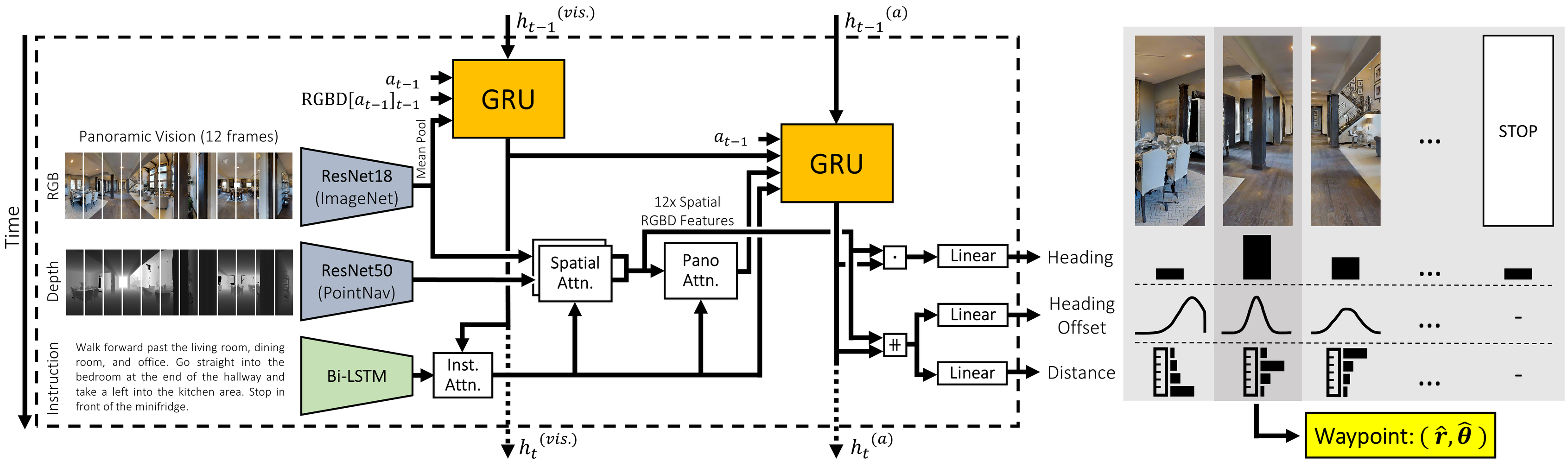}
    \hfill
    \vspace{-10pt}
    \caption{We develop a waypoint prediction network (WPN) that predicts relative waypoints directly from natural language instructions and panoramic vision. Our WPN uses two levels of cross-modal attention and prediction refinement to align visual observations with actions.}
    \label{fig:network}
\end{figure*}

An overview of our network is shown in \figref{fig:network}. At each time step, our agent observes the world through a panoramic RGBD sensor represented by 12 RGBD observations captured at regular angular intervals ($\theta$\;=\;0, 15, 30, ..., 330). Our agent predicts the next navigation waypoint in relative coordinates by selecting one of these discrete observations as a coarse heading $\hat{\theta}^D$ and then predicting an angular offset $\hat{\theta}^{offset}$ and distance $\hat{r}$ such that the waypoint is specified by the polar coordinates ($\hat{r}, \hat{\theta}^D + \hat{\theta}^{offset})$. We base our model architecture on the cross-modal attention network of Krantz \etal \cite{krantz2020beyond}, adapting the single-modality visual encoders to panoramas, adding attention over panorama frames, and developing action generation layers for waypoint prediction.

\xhdr{Visual Encoding.}
Our network encodes RGB and depth observations separately. Each RGB frame is encoded with a ResNet-18 \cite{he_cvpr16} pre-trained on ImageNet, collectively producing features $\mathcal{V}_t \in \mathbb{R}^{12 \times i \times j}$ for 12 frames containing $i$ feature map channels of flattened spatial dimensions $j$. Similarly, each depth frame is encoded with a ResNet-50 pre-trained on a PointGoal navigation task \cite{wijmans2019dd}, collectively producing features $\mathcal{D}_t \in \mathbb{R}^{12 \times k \times l}$. We provide static pose features $\mathcal{P} \in \mathbb{R}^{12 \times 2}$ consisting of the sine and cosine of the camera angle. These features disambiguate the relative angle between frames and are commonly used by VLN panorama agents to encode previous actions \cite{fried2018speaker}.

The difference between an agent's visual observations at time $t$ \vs $t-1$ can be more substantial with waypoint-based navigation than with a lower-level action space, \eg, when a waypoint is predicted through a doorway. We provide this visual context explicitly by including a subset of visual features from the previous time step. Specifically, we include features for the panorama frame facing nearest the heading of the last waypoint prediction: $\mathcal{V}_{t-1}^{(i)}$ and $\mathcal{D}_{t-1}^{(i)}$ where $i=\hat{\theta}^D_{t-1}$. These features are mean-pooled across their spatial dimension, resulting in a visual context vector $\bar{\mathcal{C}} = [\mathcal{\bar{V}}_{t-1}^{(i)}, \mathcal{\bar{D}}_{t-1}^{(i)}]$, where $[\cdot]$ denotes concatenation. 

\xhdr{Instruction Encoding.} We use the same instruction encoding as Krantz \etal \cite{krantz2020beyond}. The natural language instruction $\mathcal{O}^{\text{inst.}}$ is a lightly-tokenized sequence of words observed at each time step. We map $\mathcal{O}^{\text{inst.}}$ to a sequence of GloVE \cite{pennington_emnlp14} embeddings $w_1, w_2, ..., w_{\text{N}}$ for an instruction of length N words. A bi-directional LSTM then produces hidden states
\begin{align}
    \mathcal{S} = \left\{ s_1, s_2, ..., s_{\text{N}} \right \} = \text{BiLSTM}(w_1, w_2, ..., w_{\text{N}}).
\end{align}

\xhdr{Previous Action Encoding.}
Our network observes the predicted waypoint from the previous time step as a vector $a_{t-1} = [\hat{r}_{t-1}, \text{sin}(\hat{\theta}_{t-1}^{D}), \text{cos}(\hat{\theta}_{t-1}^{D}), \hat{\theta}_{t-1}^{offset}]$.

\xhdr{Visual History.} We use a dedicated recurrent network to track visual history like Krantz \etal \cite{krantz2020beyond}, including inputs of RGB features $\mathcal{V}_t$, the previous action $a_{t-1}$, and the additional visual context $\bar{\mathcal{C}}$. We mean-pool $\mathcal{V}_t$ across both the spatial and frame dimensions, resulting in vector $\bar{\mathcal{V}}_t$. Our visual history is then encoded as
\begin{align}
    h^{(vis.)}_t &= \text{GRU} \left (
        \left [ \bar{\mathcal{V}}_t, \bar{\mathcal{C}}, a_{t-1} \right ], h^{(vis.)}_{t-1}
    \right ).
\end{align}
\xhdr{Cross-Modal Attention.} We use scaled dot-product attention (Attn) for all attention mechanisms in our network \cite{vaswani2017attention}. The output of the visual history module $h^{(vis.)}_t$ attends to the recurrent instruction features $\mathcal{S}$:
\begin{align}
    \hat{\mathcal{S}} = \text{Attn} \left ( \mathcal{S}, h^{(vis.)}_t \right ).
\end{align}
These attended instruction features are then used to perform spatial attention on each RGB and depth frame $i$:
\begin{align}
    \hat{\mathcal{V}}_t^{(i)} = \text{Attn} \left ( \bar{\mathcal{V}}_t^{(i)}, \hat{\mathcal{S}} \right ) \;\;\;\;
    \hat{\mathcal{D}}_t^{(i)} = \text{Attn} \left ( \bar{\mathcal{D}}_t^{(i)}, \hat{\mathcal{S}} \right )
\end{align}
which is shown in \figref{fig:network} as \texttt{Spatial Attn}. The resulting features are concatenated with pose features $\mathcal{P}$, resulting in instruction-conditioned and heading-aware RGBD features for each panorama frame:
\begin{align}
    \hat{\mathcal{I}_t} = \left [ \hat{\mathcal{V}}_t, \hat{\mathcal{D}}_t, \mathcal{P} \right ].
\end{align}
The attended instruction features are used again to attend across panorama frames (\texttt{Pano Attn} in \figref{fig:network}) prior to a final recurrent block:
\begin{align}
    \hat{\mathcal{X}} &= \text{Attn} \left ( \hat{\mathcal{I}_t}, \hat{\mathcal{S}} \right ) \\
    h^{(a)}_t &= \text{GRU} \left (
        \left [ \hat{\mathcal{X}}, \hat{\mathcal{S}}, h^{(vis.)}_{t}, a_{t-1} \right ], h^{(a)}_{t-1}
    \right ).
\end{align}
\xhdr{Action Prediction.} We use the final recurrent state $h_t^{(a)}$ and the frame-specific features $\hat{\mathcal{I}}_t$ to predict a waypoint in relative polar coordinates. Our waypoint prediction begins as a coarse heading prediction sampled from a distribution over the 12 frames and a \texttt{STOP} action: $\hat{\theta}^D \sim \texttt{Pano}$. The logits of \texttt{Pano} are the dot product between $h_t^{(a)}$ and $\hat{\mathcal{I}}_t$:
\begin{align}
\texttt{Pano} = \text{softmax} \left ( \left [ \hat{\mathcal{I}_t} \cdot h^{(a)}_t, W_s h^{(a)}_t + b_s \right ] \right ).   
\end{align}

For each frame heading $i$ in \texttt{Pano}, we predict distributions over a heading offset refinement and a distance as shown in \figref{fig:network}. In \secref{sec:express}, we explore how the expressivity of the waypoint action space affects performance. To support those experiments, our offset and distance distributions are either continuous, discrete, or constant. We use the truncated Gaussian distribution \cite{burkardt2014truncated} for fixed-range continuous predictions and parameterize it by predicting the mean and variance of the underlying Gaussian:
\begin{align}
    \texttt{Offset}^{(i)} &= \text{tanh} \left (
        W_o \left [ \hat{\mathcal{I}}_t^{(i)}, h^{(a)}_t \right ] + b_o
    \right ) \text{and} \\
    \texttt{Dist}^{(i)} &= \text{sigmoid} \left (
        W_d \left [ \hat{\mathcal{I}}_t^{(i)}, h^{(a)}_t \right ] + b_d
    \right )
\end{align}
where the range of $\texttt{Offset}^{(i)}$ is $[-15\degree, 15\degree]$ and the range of $\texttt{Dist}^{(i)}$ is $[0.25\text{m}, 4.0\text{m}]$.
For discrete distributions, we replace the tanh and sigmoid activation functions with a softmax for an offset domain of $\{-15\degree, -10\degree,, ..., 15\degree \}$ and a distance domain of $\{ 0.25\text{m}, 0.75\text{m}, ..., 2.75\text{m} \}$. For constant predictions, the offset is $0\degree$ and the distance is $0.25\text{m}$, corresponding to the forward step size of the standard VLN-CE action space.

We sample a heading offset $\hat{\theta}^{offset} \sim \texttt{Offset}^{(\hat{\theta}^D)}$ and a distance $\hat{r} \sim \texttt{Dist}^{(\hat{\theta}^D)}$ conditioned on the chosen coarse heading $\hat{\theta}^D$. This produces a polar waypoint prediction $(\hat{r}, \hat{\theta}^{D} + \hat{\theta}^{offset})$. We visualize a set of possible waypoint action spaces in \tabref{tab:ablations}.

\csubsection{Training the Waypoint Prediction Network}

Existing work on the VLN-CE task trains agents with imitation learning~\cite{krantz2020beyond}. Motivated by recent advancements in embodied navigation, we instead train our waypoint prediction network with decentralized distributed proximal policy optimization (DDPPO)~\cite{wijmans2019dd}. DDPPO is a scaled version of the proximal policy optimization (PPO) algorithm with an actor-critic policy structure~\cite{schulman2017proximal}. We consider the loss function used in~\cite{wijmans2019dd} for PointGoal navigation. It employs the clipped PPO objective $L_\text{action}$, a clipped critic loss $L_\text{value}$, and an entropy bonus $L_S$ to encourage exploration:
\begin{align}
    L_{\text{standard}} = L_\text{action} + c_v L_\text{value} - c_e  L_S.
\end{align}
Let $\theta$-parameterized policy $\pi_\theta$ be the waypoint prediction network. For $L_\text{action}$, we compute the probability $\pi_\theta(\mathcal{A}_t)$ of an action $\mathcal{A}_t = (\hat{\theta}^{D}, \hat{\theta}^{offset}, \hat{r})$ for a panorama frame selection $\hat{\theta}^{D}$, a heading offset $\hat{\theta}^{offset}$, and a distance $\hat{r}$ as:
\begin{align}
	\texttt{Pano}(\hat{\theta}^{D}&) * \texttt{Offset}^{(\hat{\theta}^{D})}(\hat{\theta}^{offset}) * \texttt{Dist}^{(\hat{\theta}^{D})}(\hat{r}). 
\end{align}
Accordingly, we define the entropy term $L_S$ as:
\begin{align}
    L_S = c_{p} S \left ( \texttt{Pano} \right ) + c_{o} S \left ( \texttt{Offset} \right ) + c_{d} S \left ( \texttt{Dist} \right )
\end{align}
to control the amount of exploration within specific action components. For $L_\text{value}$, we predict a state-value estimate from the final hidden state $h_t^{(a)}$ as $\hat{v} = \text{linear}(h_t^{(a)})$.

We expand this loss function with an additional zero-trending regularization term  $L_\text{offset} = \left |\hat{\theta}^{offset} \right|$, which we found empirically led to better exploration of the joint \texttt{Pano}-\texttt{Offset} heading space. Together, this yields our total loss function:
\begin{align}
L_\text{total} &= L_{\text{standard}} + c_r  L_\text{offset}. 
\end{align}

\setlength{\tabcolsep}{.3em}
\begin{table*}[t]
	\begin{center}
	\resizebox{0.975\textwidth}{!}{
	    \begin{minipage}{0.34\textwidth}
	    \includegraphics[width=\textwidth]{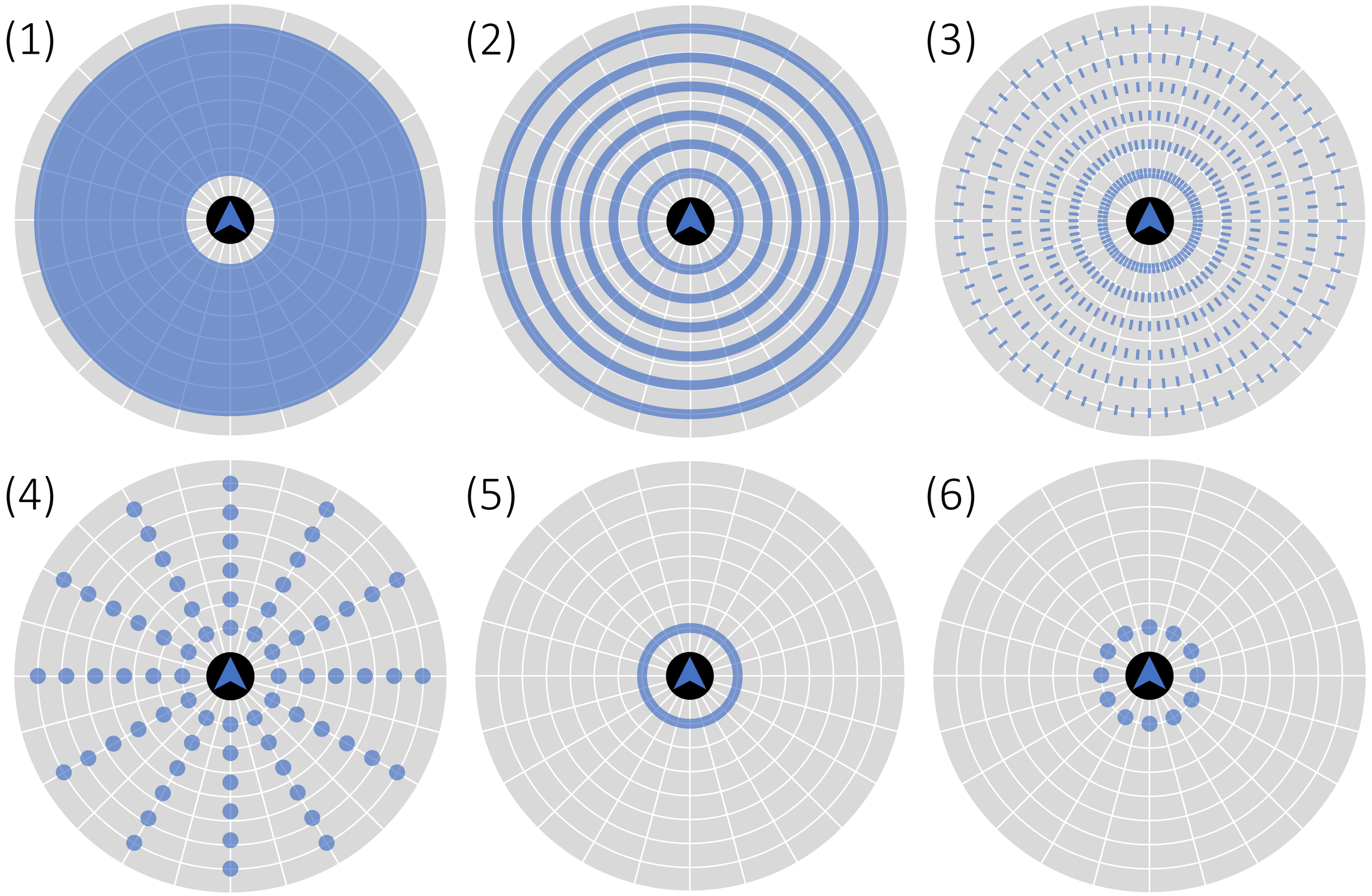}
        \end{minipage}
        \hfill
        \begin{minipage}{0.65\textwidth}
        \resizebox{\textwidth}{!}{
		\begin{tabular}{l l cc c cccsc c cccsccc}
			\toprule

			&& \multirow{2}{*}[-1em]{\scriptsize \shortstack{Dist.}}&\multirow{2}{*}[-1em]{\scriptsize \shortstack{Offset}} & & \multicolumn{5}{c}{\scriptsize\textbf{Val-Seen}} & & \multicolumn{6}{c}{\scriptsize\textbf{Val-Unseen}}  \\
			\cmidrule{6-10} \cmidrule{12-18}
			\scriptsize \texttt{\#} &\scriptsize Model &    &  & & \scriptsize\textbf{\texttt{TL}}~ & \scriptsize\textbf{\texttt{NE}}~$\downarrow$ & \scriptsize\textbf{\texttt{OS}}~$\uparrow$ & \scriptsize\textbf{\texttt{SR}}~$\uparrow$ &\scriptsize\textbf{\texttt{SPL}}~$\uparrow$ & & \scriptsize\textbf{\texttt{TL}}~ & \scriptsize\textbf{\texttt{NE}}~$\downarrow$ &
			\scriptsize\textbf{\texttt{OS}}~$\uparrow$ & \scriptsize\textbf{\texttt{SR}}~$\uparrow$ & \scriptsize\textbf{\texttt{SPL}}~$\uparrow$ & \scriptsize\textbf{\texttt{EET}} & \scriptsize\textbf{\texttt{SCT}}~$\uparrow$\\
			\midrule

			\scriptsize \texttt{1} & \multirow{4}{*}{\small \shortstack[l]{Waypoint Pred.\\ Network (WPN)}} & C & C & & 10.29 & 6.05 & 51 & 40 & 35 &   & 10.38 & 6.90 & 41 & 34 & 29 & 186 & 20\\
			\scriptsize \texttt{2} && D & C & & 10.51 & 6.12 & 49 & 38 & 33 &   & 10.62 & 6.62 & \textbf{43} & 36 & 30 & 153 & \textbf{23}\\
			 \scriptsize \texttt{3} & & D & D & & 9.11 & 6.57 & 44 & 35 & 32 &   & 8.23 & 7.48 & 35 & 28 & 26 & 93 & 20\\
			 \scriptsize \texttt{4} & & D & - & & 9.06 & 6.45 & 46 & 39 & 35 &   & 8.16 & 7.20 & 38 & 31 & 28 & 90 & 22\\
			\midrule
			\scriptsize \texttt{5} &\multirow{2}{*}{\small \shortstack[l]{ Heading Pred.\\Network (HPN)}} & - & C & & 8.71 & \textbf{5.17} & 53 & \textbf{47} & \textbf{45}   &  & 7.71 & \textbf{6.02} & 42 & 38 & \textbf{36}  & 297 & 11\\
			\scriptsize \texttt{6} & & - & - && 8.63 & 5.44 & 51 & 44 & 42  &  & 7.72 & 6.21 & 38 & 34 & 32 & 308 & 11\\
			\bottomrule
		\end{tabular}}
		\end{minipage}}
	\end{center}
	\caption{Results of our waypoint model on Val-Seen and Val-Unseen splits using the continuous navigator to reach waypoints. We demonstrate the influence of our action space components by successively constraining the waypoint action space. We find that the least-constrained heading prediction network performs the best according to conventional VLN metrics across both validation splits.}
	\vspace{-8pt}
	\label{tab:ablations}
\end{table*}

\xhdr{Reward Function.} Our reward function is informed by the extrinsic reward structure of Wang \etal \cite{wang2019reinforced} and the time penalty (or slack reward) from Savva \etal \cite{habitat19arxiv}. We include a success reward $r_\text{success}$, the change in distance to target $\Delta_{\text{dist\_to\_target}}$ and a slack reward $r_\text{slack}$:
\begin{align}
r(s, t) & = r_\text{success} - \Delta_{\text{dist\_to\_target}} + r_\text{slack},
\end{align}
where $r_\text{success}=2.5$ once a stop action is called within $3\text{m}$ of the target location (otherwise equal to $0$) and $\Delta_{\text{dist\_to\_target}} = D(s_{t}) - D(s_{t-1})$ is progress towards the goal in terms of geodesic distance. The slack reward as defined by~\cite{habitat19arxiv} is constant and applied at every time step. A waypoint predictor that maximizes this reward term would predict the furthest navigable waypoint toward the goal. This is undesirable for instruction-following where agents need to consider intermediate navigation decisions in light of partial observability. In the instruction \textit{``go into the bedroom"}, an agent must first decide to continue past other similar-looking doorways (such as a bathroom) before choosing to enter the bedroom. We mitigate this training bias toward distant waypoints by scaling the slack reward based on waypoint distance instead of time. Specifically, we scale slack based on distance predicted: $r_\text{slack} = - 0.05 \cdot \frac{ d_{predicted}}{0.25m}$ which additionally penalizes unreachable waypoints.

\csection{Experiments}

\noindent In this section, our main experiments address the following questions within the context of the VLN-CE task: 

\xhdr{1) How does the expressivity of waypoint predictions affect performance?} On one end of the expressivity spectrum, an agent may select waypoints from a small set of discrete candidates, and on the other end, an agent may consider any continuous location within some range. We examine the impact of different levels of expressivity in \secref{sec:express}. Generally, we find that less expressive action spaces lead to minor improvements in standard metrics over more expressive versions but result in trajectories that would be slower to execute on real agents due to frequent stops and turns.

\xhdr{2) How do our waypoint-based models compare to prior work in low-level action spaces?} Compared to existing  work on VLN-CE~\cite{krantz2020beyond}, our base models are trained with additional sensors (forward-facing \vs ~panoramic cameras) and continuous navigators with arbitrary turn angles and step distances. While we argue these observation and navigator action spaces are \emph{more reflective} of real robotic agents, we ablate these in \secref{sec:compare} to compare with prior work. We find that our models result in significant improvements over prior work on the public VLN-CE leaderboard.

\csubsection{Experimental Setup}

\xhdrflat{VLN-CE Dataset.} We use the VLN-CE dataset \cite{krantz2020beyond} which consists of $16,844$ path-instruction pairs (5,611 unique paths) across 90 scenes. The dataset is split into train (Train), seen validation (Val-Seen), unseen validation (Val-Unseen), and test (Test). Both Val-Unseen and Test contain scenes the agent has not been exposed to during training.

\xhdr{Metrics.} We evaluate our agent using established metrics from VLN-CE \cite{krantz2020beyond}. Specifically, we report the metrics used by the VLN-CE Challenge leaderboard which include trajectory length (\texttt{TL}), navigation error (\texttt{NE}), oracle success rate (\texttt{OS}), success rate (\texttt{SR}), and success weighted by inverse path length (\texttt{SPL}). Note that success occurs when an agent invokes the \texttt{stop} action within 3m of the goal. Please see \cite{anderson2018evaluation, anderson2018vision} for a detailed description of these metrics. 

\xhdr{Implementation Details.}
We implement our agents in PyTorch \cite{paszke2019pytorch} and use the Habitat Simulator \cite{habitat19arxiv}. We extend Habitat's DDPPO \cite{wijmans2019dd} training implementation to the VLN-CE task and add components for training waypoint prediction agents. We distribute training across 64 GPUs, collecting around 200M steps of experience to reach peak performance (5 days on average). We use the same set of hyperparameters for each experiment and include those values in the supplementary. We use early stopping during the training process and select the checkpoint with the highest \texttt{SPL} on Val-Unseen for all models. During the evaluation, the waypoint prediction network takes the mode of each action distribution which leads to deterministic results.

\setlength{\tabcolsep}{.3em}
\begin{table*}[t]
	\begin{center}
	\resizebox{0.975\textwidth}{!}{
	\begin{minipage}{0.38\textwidth}
	\includegraphics[width=\textwidth]{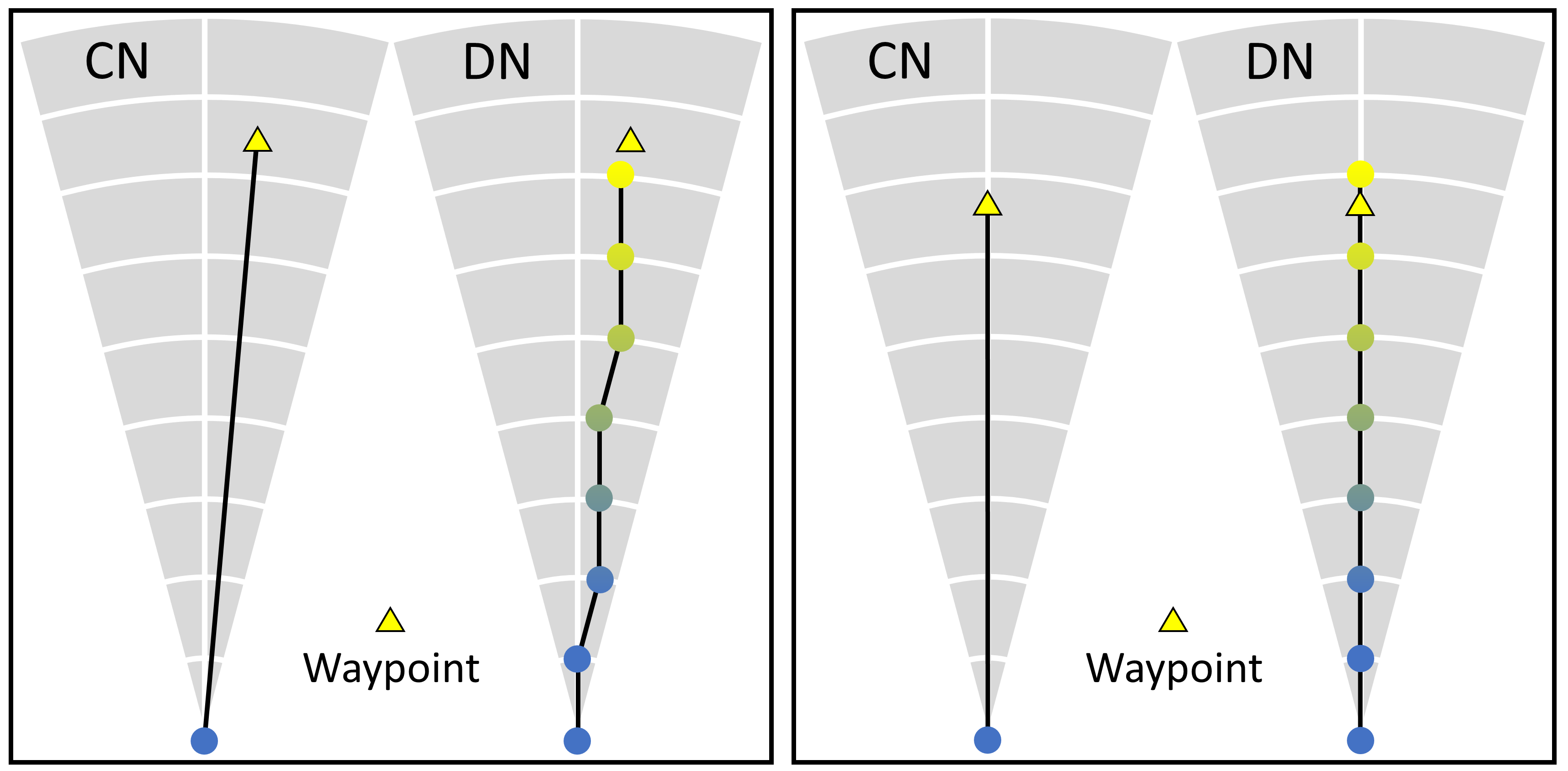}
	\end{minipage}
	\hfill
	\begin{minipage}{0.61\textwidth}
	\resizebox{\textwidth}{!}{
		\begin{tabular}{l l cc c cccsc c cccsc}
			\toprule

			&& \multirow{2}{*}[-1em]{\scriptsize \shortstack{Navigator}}&\multirow{2}{*}[-1em]{\scriptsize \shortstack{Ckpt}} & & \multicolumn{5}{c}{\scriptsize\textbf{Val-Seen}} & & \multicolumn{5}{c}{\scriptsize\textbf{Val-Unseen}}  \\
			\cmidrule{7-10} \cmidrule{12-16}
			\scriptsize \texttt{\#} &\scriptsize Model &    & & & \scriptsize\textbf{\texttt{TL}}~$\downarrow$ & \scriptsize\textbf{\texttt{NE}}~$\downarrow$ & \scriptsize\textbf{\texttt{OS}}~$\uparrow$ & \scriptsize\textbf{\texttt{SR}}~$\uparrow$ &\scriptsize\textbf{\texttt{SPL}}~$\uparrow$ & & \scriptsize\textbf{\texttt{TL}}~$\downarrow$ & \scriptsize\textbf{\texttt{NE}}~$\downarrow$ &
			\scriptsize\textbf{\texttt{OS}}~$\uparrow$ & \scriptsize\textbf{\texttt{SR}}~$\uparrow$ & \scriptsize\textbf{\texttt{SPL}}~$\uparrow$\\
			\midrule

			\scriptsize \texttt{1} & \multirow{3}{*}{\shortstack[l]{WPN,\\discrete\\distance}} & CN & 222 && 10.51 & 6.12 & 49 & 38 & 33 &   & 10.62 & 6.62 & 43 & 36 & 30 \\
			 \scriptsize \texttt{2} & & DN & 222 && 9.64 & 6.33 & 43 & 34 & 30 &   & 9.54 & 6.85 & 40 & 33 & 28 \\
			 \scriptsize \texttt{3} && DN & 89 && 9.52 & 6.23 & 45 & 37 & 33 &   & 9.86 & 6.93 & 40 & 33 & 29 \\
			\midrule
			\scriptsize \texttt{4} &\multirow{3}{*}{ \shortstack[l]{WPN,\\continuous\\distance}} & CN & 137 && 10.29 & 6.05 & 51 & 40 & 35   &  & 10.38 & 6.90 & 41 & 34 & 29  \\
			\scriptsize \texttt{5} & & DN & 137 && 10.14 & 5.99 & 52 & 42 & 36  &  & 9.60 & 6.87 & 39 & 32 & 28 \\
			\scriptsize \texttt{6} & & DN & 185 && 10.73 & 5.99 & 52 & 41 & 36 &  & 10.61 & 7.07 & 42 & 33 & 28 \\
			\bottomrule
		\end{tabular}}
	\end{minipage}}
	\end{center}
	\caption{Validation performance of our waypoint prediction network (WPN) paired with different navigators. Despite training with a continuous navigator (CN), our WPN drops only 1-2 \texttt{SPL} in Val-Unseen using a discrete navigator (DN).}
	\label{tab:navigators}
\end{table*}

\csubsection{Impact of Waypoint Expressivity}

\label{sec:express}
To study the effect of waypoint expressivity, we consider a spectrum of prediction domains for our model's distance and offset components. In \tabref{tab:ablations}, we consider predicting continuous values (\texttt{C}), choosing between a set of discrete values (\texttt{D}), or not predicting at all and using a fixed value (\texttt{-}). These combinations result in decision spaces visualized in the figure on the left of \tabref{tab:ablations} where blue-shaded regions reflect possible waypoints under various \texttt{C}/\texttt{D}/\texttt{-} settings of offset and distance prediction. The labels at the top-left of each graph match the corresponding row(s) of the table.

\xhdr{WPN.} Row 1 is our fully continuous waypoint prediction network (WPN) which can select any point within a toroid around the agent bounded by 0.25 and 4m. In row 2, we consider discrete distance prediction over six choices ranging from 0.25m to 2.75m in increments of 0.5m -- resulting in a decision space of six continuous rings. In row 3, we additionally constrain the offset to seven choices ranging from $-15\degree$ to $15\degree$ in increments of $5\degree$ -- further segmenting the rings into a dense set of discrete points. In row 4, we fix the offset prediction to 0, resulting in a sparse `wagon-wheel' of 36 possible waypoints. This progresses from fully continuous to highly constrained subspaces.

We observe the most significant differences in performance from changes to the offset prediction space. Continuous offsets outperform their discrete or fixed counterparts by 3-8\% success (rows 1 \& 2 \vs 3 \& 4). Intuitively, continuous offset prediction enables more position control at longer distances (compare the outer edge of plot 1 with 4). Surprisingly the dense discrete setting (row 3) under-performed no offsets (row 3 \vs 4) by 3\% success. We suspect this is due to differences in training dynamics -- we observe rapid training convergence for this model which could lead to relative under-exploration of the action space.

\begin{table*}[t]
	\setlength{\tabcolsep}{3pt}
	\begin{center}
		\begin{tabular}{c l c cccsc c cccsc c cccsc}
			\toprule

			& & & \multicolumn{5}{c}{\scriptsize\textbf{Val-Seen}} & & \multicolumn{5}{c}{\scriptsize\textbf{Val-Unseen}}  & & \multicolumn{5}{c}{\scriptsize\textbf{Test}}  \\
			\cmidrule{4-8} \cmidrule{10-14} \cmidrule{16-20} 
			\scriptsize \texttt{\#} & \scriptsize Model & & \scriptsize\textbf{\texttt{TL}}~$\downarrow$ & \scriptsize\textbf{\texttt{NE}}~$\downarrow$ & \scriptsize\textbf{\texttt{OS}}~$\uparrow$ & \scriptsize\textbf{\texttt{SR}}~$\uparrow$ &\scriptsize\textbf{\texttt{SPL}}~$\uparrow$ & & \scriptsize\textbf{\texttt{TL}}~$\downarrow$ & \scriptsize\textbf{\texttt{NE}}~$\downarrow$ & \scriptsize\textbf{\texttt{OS}}~$\uparrow$ & \scriptsize\textbf{\texttt{SR}}~$\uparrow$ & \scriptsize\textbf{\texttt{SPL}}~$\uparrow$ & & \scriptsize\textbf{\texttt{TL}}~$\downarrow$ & \scriptsize\textbf{\texttt{NE}}~$\downarrow$ & \scriptsize\textbf{\texttt{OS}}~$\uparrow$ & \scriptsize\textbf{\texttt{SR}}~$\uparrow$ &\scriptsize\textbf{\texttt{SPL}}~$\uparrow$\\
			\midrule
			\scriptsize \texttt{1} & HPN + DN (ours) && 8.54 & \textbf{5.48} & \textbf{53} & \textbf{46} & \textbf{43} &  & 7.62 & \textbf{6.31} & \textbf{40} & \textbf{36} & \textbf{34} &   & 8.02 & \textbf{6.65} & \textbf{37} & \textbf{32} & \textbf{30}\\
			\scriptsize \texttt{2} & WPN + DN (ours) && 9.52 & 6.23 & 45 & 37 & 33 &  & 9.86 & 6.93 & \textbf{40} & 33 & 29 &   & 9.68 & 7.49 & 36 & 29 & 25\\
			\scriptsize \texttt{3} & CMA+PM+DA+Aug \cite{krantz2020beyond} && 9.06 & 7.21 & 44 & 34 & 32 &  & 8.27 & 7.60 & 36 & 29 & 27 &   & 8.85 & 7.91 & 36 & 28 & 25\\
            \bottomrule
		\end{tabular}
	\end{center}
	\caption{Results on the VLN-CE Challenge leaderboard. Both of our model submissions outperform existing state of the art on Test, with our heading prediction network (HPN) showing the highest success rate (\texttt{SR}) with the lowest trajectory length (\texttt{TL}).}\vspace{-8pt}
	\label{tab:leaderboard}
\end{table*}

\begin{figure*}[t]
    \centering
    \includegraphics[width=\textwidth]{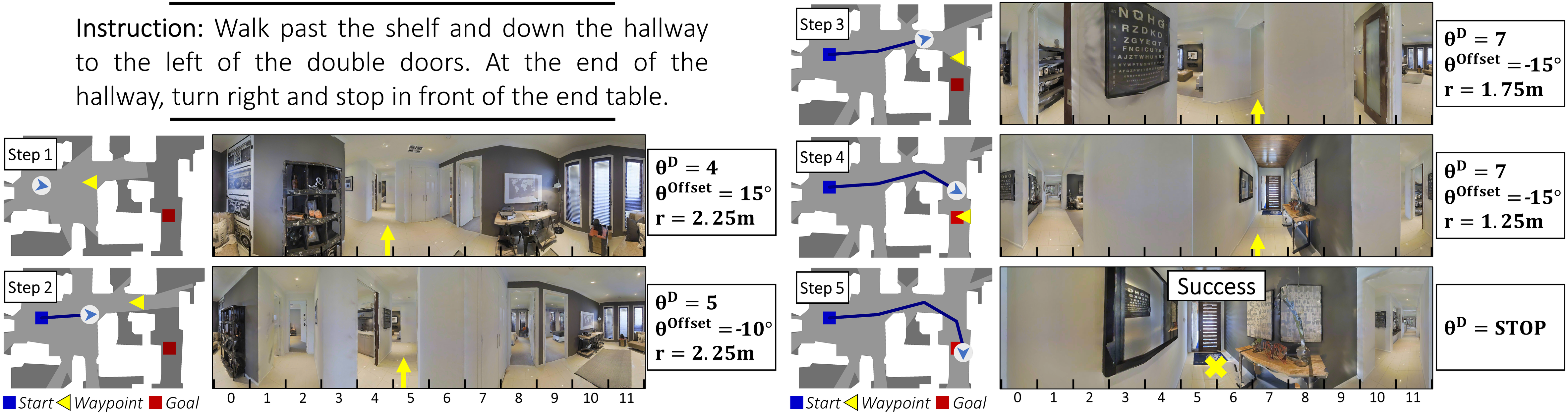}
    \caption{A qualitative example of our best waypoint agent (WPN+CN) successfully navigating to the goal in an unseen environment.}\vspace{-8pt}
    \label{fig:qualitative}
\end{figure*}

\xhdr{HPN.}
In rows 5 \& 6 we ablate distance prediction entirely, moving a fixed 0.25m in the chosen heading. To reflect this, we refer to these ablations as Heading Prediction Networks (HPNs). For a continuous offset (row 5), this allows a waypoint to be predicted in a single ring of radius 0.25m. Row 6 further ablates the offset prediction, resulting in a ``pick-pano" model that effectively mimics the existing VLN-CE action space \texttt{Forward-Left-Right} with collapsed turn actions (e.g.~reducing any consecutive sequence of turns followed by a forward step into a single action). As before, we observe continuous offsets lead to improvements. 

Counter-intuitively, we find these fixed-distance models generally outperform their WPN counterparts in terms of success by 2-3\% (e.g.~ rows 1/2 \vs  5 and row 3 \vs 6). However, these agents take approximately 4x the actions per trajectory (row 2 at 8.41 and row 5 at 33.41)-- resulting in paths with significantly more starts, stops, and turns. Consequently, these high action-rate paths more closely approximate the ground truth path and achieve higher path efficiency as shown in \texttt{SPL}. In contrast, the WPN models break the path down into straight-line segments of 1.6 meters on average -- reducing time to execute on real systems.

Given the variance associated with RL training methods, we repeat the experiment in row 2 of \tabref{tab:ablations} twice under different random seeds. Both achieve a 29 \texttt{SPL} in Val-Unseen (1 point lower than row 2), suggesting that performance differences of 1 \texttt{SPL} may not be significant.

\xhdr{Path Efficiency under a LoCoBot Motion Model.}  Depending on a robot's abilities, shorter-length paths with many fine-grained actions may take considerably longer to execute than simpler-but-longer ones. We profile a LoCoBot~\cite{gupta2018robot} robot controlled via PyRobot~\cite{pyrobot2019}. We choose LoCoBot because it is a common platform for sim2real experiments in embodied tasks~\cite{kadian_sim2realgap_2020, deitke2020robothor, chaplot2020learning}. We derive functions for the time to turn by a specified angle or move forward by a specified distance from empirical measurements. Using these, we can estimate the time a LoCoBot would take for any path. For more details, see the supplementary.

We call this metric the estimated execution time (\texttt{EET}) and present results for each model in unseen environments in \tabref{tab:ablations}. We report \texttt{EET} in seconds. Intuitively, we find that models that predict travel distance (rows 1-4) have a lower \texttt{EET} than models that step in fixed $0.25$m increments (rows 5-6). In particular, our best WPN (row 2) has a lower \texttt{EET} than our best HPN (row 5) by 144 seconds---nearly a 2x reduction. Digging into this further, we can compare the estimated average speed during a trajectory by normalizing trajectory length by \texttt{EET} (\texttt{TL}/\texttt{EET}). Our best WPN averages 6.9 cm/s, a 2.7x increase over our best HPN at just 2.6 cm/s.

We additionally present success weighted by completion time (\texttt{SCT}) \cite{yokoyama2021success} which scales the agent's success by the relative time to complete the trajectory. We adapt \texttt{SCT} to our agent's dynamics by using \texttt{EET} for completion times. Details are in the supplementary. We find that our best WPN model has over a 2x improvement in \texttt{SCT} over our best HPN model (23 \vs 11) despite WPN having a lower \texttt{SPL}. These results demonstrate the practical benefit of using waypoint models for real-world execution.

\csubsection{Comparison with Discrete Action Models}
\label{sec:compare}
Our agents are trained with continuous navigators that can turn to arbitrary angles and move forward by arbitrary distances -- matching realistic zero-turn radius robots. In contrast, VLN-CE assumes turns of 15 degree increments and forward steps of 0.25m. To compare with prior work, we implement a discrete navigator (DN) that uses this low-level action space to reach waypoints approximately. Our DN assumes free space and selects actions that greedily minimize distance to the waypoint. We assume no explicit localization. \tabref{tab:navigators} shows our WPN model using continuous \vs discrete navigators at inference. As shown in the figure (left), the discrete navigator approximates the path of the continuous version. We find our models are somewhat robust to this change in navigator but drop 1-3\% success. In rows 3 \& 6 we re-evaluate all model checkpoints using the discrete navigator, finding that while different checkpoints maximize \texttt{SPL}, the performance is similar to rows 2 \& 5.

In \tabref{tab:leaderboard}, we compare our models using a discrete navigator (DN) with prior work on the VLN-CE leaderboard. We submit both our best performing waypoint prediction network (WPN) and heading prediction network (HPN) variants based on Val-Unseen \texttt{SPL}. The existing state-of-the-art belongs to a cross-modal attention model trained by dataset aggregation (DAgger) and aided by progress monitor and data augmentation (CMA+PM+DA+Aug) \cite{krantz2020beyond}. Both WPN+DN and HPN+DN surpass the performance of existing work, with HPN+DN setting the new state of the art on the VLN-CE task by 4 \texttt{SR} (14\% relative) and 5 \texttt{SPL} (20\% relative). This  is despite evaluating our networks with a navigator they were not exposed to during training.

Looking closely at the differences between prior work and our HPN+DN model, our agent has access to panoramic observations, has a more abstract heading-based action space, and is trained with RL. To ablate these differences, we start from our ``pick-pano" HPN model (\tabref{tab:ablations} row 6) and ablate panoramic observation to a single forward-facing camera. In Val-Unseen, this model surpasses row 6 by 2 \texttt{SR} (achieving a 36/32 \texttt{SR}/\texttt{SPL}). This agent has a lower seen-to-unseen gap than row 6 by 4 \texttt{SR} and 5 \texttt{SPL} (Val-Seen: 40/37 \texttt{SR}/\texttt{SPL}). This suggests that the reduced visual information of this model leads to less overfitting of the training environments. We further ablate the high-level action space of this model, matching the observation and action spaces of prior work. We find that this agent is unable to train to convergence after 300M steps of experience and identify the longer time horizon as a challenge requiring deeper study.

\csubsection{Qualitative Example}

We present a qualitative example of our best waypoint agent navigating an unseen environment (\figref{fig:qualitative}). In Step 1, the agent traverses a large room by predicting a waypoint $2.25$m away. In Step 4, the waypoint prediction is shorter at $1.25$m, directly in front of the end table referenced in the instruction. Together, these predictions demonstrate the agent's ability to implicitly reason about scene geometry and predict language-grounded waypoints. Each step in this example can be aligned with an abstract semantic sub-goal, \eg ``\textit{continue through the hallway}" (Step 2) and ``\textit{go to the end table}" (Step 4). Agents that directly predict actions from the VLN-CE action space need to make 10+ decisions to execute each sub-goal -- an unintuitive and time-inefficient exercise. We provide additional navigation examples in the supplementary.

\csubsection{Waypoint Prediction Analysis}

We analyze characteristics of the waypoints predicted by our best WPN model (\tabref{tab:ablations} row 2). In both Val-Seen and Val-Unseen, the mean distance prediction is 1.6m with a standard deviation of 0.8m. We find that waypoint distances decrease with time, such that that predictions in the first 25\% of an episode average 2.3m, predictions in the middle 50\% average 1.6m, and predictions in the final 25\% average 0.76m. This behavior is reasonable in the context of instruction-following -- commonly, the beginning of a path is described as taking macro actions (\eg ``\textit{Exit the bedroom...}"), while the end of a path can be described more particularly (\eg ``\textit{...and wait between the two chairs.}").

\csection{Discussion}
In this work, we present a model class that predicts navigation waypoints directly from language and vision. In exploring the expressivity of the waypoint action space, we find that more expressive models have favorable real-world execution properties, including a 2x reduction in expected execution time and a modular architecture that abstracts interaction with robot-specific navigation stacks. On the other hand, less expressive action spaces lead to higher traditional VLN metrics. Our best submission to the VLN-CE leaderboard demonstrates this through a 4\% improvement in success (14\% relative) and a 5 point improvement in \texttt{SPL} (20\% relative) over prior work. We recognize that a significant gap still remains between topological VLN and continuous VLN-CE. Addressing this gap and the related sim2real gap \cite{anderson2020sim} will require developing an effective interface between language understanding and robotic control.

{
\small
\xhdr{Acknowledgements}
We would like to thank Naoki Yokoyama for helping adapt \texttt{SCT} and Joanne Truong for help with physical LoCoBot profiling. This work is funded in part by DARPA MCS. 
}

{\small
\bibliographystyle{ieee_fullname}
\bibliography{bib/strings,bib/main}

\begin{thebibliography}{10}\itemsep=-1pt

\bibitem{locobot}
Locobot: an open source low cost robot.
\newblock 2019.

\bibitem{anderson2018evaluation}
Peter Anderson, Angel Chang, Devendra~Singh Chaplot, Alexey Dosovitskiy,
  Saurabh Gupta, Vladlen Koltun, Jana Kosecka, Jitendra Malik, Roozbeh
  Mottaghi, Manolis Savva, et~al.
\newblock On evaluation of embodied navigation agents.
\newblock {\em arXiv preprint arXiv:1807.06757}, 2018.

\bibitem{anderson2020sim}
Peter Anderson, Ayush Shrivastava, Joanne Truong, Arjun Majumdar, Devi Parikh,
  Dhruv Batra, and Stefan Lee.
\newblock Sim-to-real transfer for vision-and-language navigation.
\newblock In {\em CoRL}, 2020.

\bibitem{anderson2018vision}
Peter Anderson, Qi Wu, Damien Teney, Jake Bruce, Mark Johnson, Niko
  S{\"u}nderhauf, Ian Reid, Stephen Gould, and Anton van~den Hengel.
\newblock Vision-and-language navigation: Interpreting visually-grounded
  navigation instructions in real environments.
\newblock In {\em CVPR}, 2018.

\bibitem{bansal2020combining}
Somil Bansal, Varun Tolani, Saurabh Gupta, Jitendra Malik, and Claire Tomlin.
\newblock Combining optimal control and learning for visual navigation in novel
  environments.
\newblock In {\em CoRL}, pages 420--429, 2020.

\bibitem{blukis2018following}
Valts Blukis, Nataly Brukhim, Andrew Bennett, Ross~A Knepper, and Yoav Artzi.
\newblock Following high-level navigation instructions on a simulated
  quadcopter with imitation learning.
\newblock In {\em RSS}, 2018.

\bibitem{blukis2018mapping}
Valts Blukis, Dipendra Misra, Ross~A Knepper, and Yoav Artzi.
\newblock Mapping navigation instructions to continuous control actions with
  position-visitation prediction.
\newblock In {\em CoRL}, 2018.

\bibitem{blukis2020learning}
Valts Blukis, Yannick Terme, Eyvind Niklasson, Ross~A Knepper, and Yoav Artzi.
\newblock Learning to map natural language instructions to physical quadcopter
  control using simulated flight.
\newblock In {\em CoRL}, pages 1415--1438, 2020.

\bibitem{burkardt2014truncated}
John Burkardt.
\newblock The truncated normal distribution.
\newblock {\em Department of Scientific Computing Website, Florida State
  University}, pages 1--35, 2014.

\bibitem{Matterport3D}
Angel Chang, Angela Dai, Thomas Funkhouser, Maciej Halber, Matthias Niessner,
  Manolis Savva, Shuran Song, Andy Zeng, and Yinda Zhang.
\newblock {Matterport3d}: learning from rgb-d data in indoor environments.
\newblock In {\em 3DV}, 2017.
\newblock {MatterPort3D} dataset license available at:
  \url{http://kaldir.vc.in.tum.de/matterport/MP_TOS.pdf}.

\bibitem{chaplot2020learning}
Devendra~Singh Chaplot, Dhiraj Gandhi, Saurabh Gupta, Abhinav Gupta, and Ruslan
  Salakhutdinov.
\newblock Learning to explore using active neural slam.
\newblock In {\em ICLR}, 2020.

\bibitem{chaplot2020object}
Devendra~Singh Chaplot, Dhiraj~Prakashchand Gandhi, Abhinav Gupta, and Russ~R
  Salakhutdinov.
\newblock Object goal navigation using goal-oriented semantic exploration.
\newblock {\em NeurIPS}, 2020.

\bibitem{chaplot2020neural}
Devendra~Singh Chaplot, Ruslan Salakhutdinov, Abhinav Gupta, and Saurabh Gupta.
\newblock Neural topological slam for visual navigation.
\newblock In {\em CVPR}, pages 12875--12884, 2020.

\bibitem{chen2020waypoints}
Changan Chen, Sagnik Majumder, Al-Halah Ziad, Ruohan Gao, Santhosh
  Kumar~Ramakrishnan, and Kristen Grauman.
\newblock Learning to set waypoints for audio-visual navigation.
\newblock In {\em ICLR}, 2021.

\bibitem{chen2019touchdown}
Howard Chen, Alane Suhr, Dipendra Misra, Noah Snavely, and Yoav Artzi.
\newblock Touchdown: Natural language navigation and spatial reasoning in
  visual street environments.
\newblock In {\em CVPR}, 2019.

\bibitem{eqamodular_corl_2018}
Abhishek Das, Georgia Gkioxari, Stefan Lee, Devi Parikh, and Dhruv Batra.
\newblock Neural modular control for embodied question answering.
\newblock In {\em CoRL}, 2018.

\bibitem{deitke2020robothor}
Matt Deitke, Winson Han, Alvaro Herrasti, Aniruddha Kembhavi, Eric Kolve,
  Roozbeh Mottaghi, Jordi Salvador, Dustin Schwenk, Eli VanderBilt, Matthew
  Wallingford, et~al.
\newblock Robothor: an open simulation-to-real embodied ai platform.
\newblock In {\em CVPR}, pages 3164--3174, 2020.

\bibitem{fried2018speaker}
Daniel Fried, Ronghang Hu, Volkan Cirik, Anna Rohrbach, Jacob Andreas,
  Louis-Philippe Morency, Taylor Berg-Kirkpatrick, Kate Saenko, Dan Klein, and
  Trevor Darrell.
\newblock Speaker-follower models for vision-and-language navigation.
\newblock In {\em NeurIPS}, 2018.

\bibitem{gonzalez2015review}
David Gonz{\'a}lez, Joshu{\'e} P{\'e}rez, Vicente Milan{\'e}s, and Fawzi
  Nashashibi.
\newblock A review of motion planning techniques for automated vehicles.
\newblock {\em T-ITS}, 17(4):1135--1145, 2015.

\bibitem{gupta2018robot}
Abhinav Gupta, Adithyavairavan Murali, Dhiraj~Prakashchand Gandhi, and Lerrel
  Pinto.
\newblock Robot learning in homes: improving generalization and reducing
  dataset bias.
\newblock {\em NeurIPS}, pages 9094--9104, 2018.

\bibitem{he_cvpr16}
Kaiming He, Xiangyu Zhang, Shaoqing Ren, and Jian Sun.
\newblock {Deep residual learning for image recognition}.
\newblock In {\em CVPR}, 2016.

\bibitem{hermann2019learning}
Karl~Moritz Hermann, Mateusz Malinowski, Piotr Mirowski, Andras Banki-Horvath,
  Keith Anderson, and Raia Hadsell.
\newblock Learning to follow directions in street view.
\newblock {\em AAAI}, 2020.

\bibitem{irshad2021hierarchical}
Muhammad~Zubair Irshad, Chih-Yao Ma, and Zsolt Kira.
\newblock Hierarchical cross-modal agent for robotics vision-and-language
  navigation.
\newblock In {\em Proceedings of the IEEE International Conference on Robotics
  and Automation (ICRA)}, 2021.

\bibitem{kadian_sim2realgap_2020}
Abhishek Kadian, Joanne Truong, Aaron Gokaslan, Alexander Clegg, Erik Wijmans,
  Stefan Lee, Manolis Savva, Sonia Chernova, and Dhruv Batra.
\newblock Are we making real progress in simulated environments? measuring the
  sim2real gap in embodied visual navigation.
\newblock In {\em IROS}, 2020.

\bibitem{krantz2020beyond}
Jacob Krantz, Erik Wijmans, Arjun Majumdar, Dhruv Batra, and Stefan Lee.
\newblock Beyond the nav-graph: Vision-and-language navigation in continuous
  environments.
\newblock In {\em ECCV}, pages 104--120, 2020.

\bibitem{ku2020room}
Alexander Ku, Peter Anderson, Roma Patel, Eugene Ie, and Jason Baldridge.
\newblock Room-across-room: multilingual vision-and-language navigation with
  dense spatiotemporal grounding.
\newblock In {\em EMNLP}, pages 4392--4412, 2020.

\bibitem{majumdar2020improving}
Arjun Majumdar, Ayush Shrivastava, Stefan Lee, Peter Anderson, Devi Parikh, and
  Dhruv Batra.
\newblock Improving vision-and-language navigation with image-text pairs from
  the web.
\newblock In {\em ECCV}, pages 259--274, 2020.

\bibitem{habitat19arxiv}
{Manolis Savva*}, {Abhishek Kadian*}, {Oleksandr Maksymets*}, Yili Zhao, Erik
  Wijmans, Bhavana Jain, Julian Straub, Jia Liu, Vladlen Koltun, Jitendra
  Malik, Devi Parikh, and Dhruv Batra.
\newblock Habitat: a platform for embodied ai research.
\newblock {\em ICCV}, 2019.

\bibitem{misra2018mapping}
Dipendra Misra, Andrew Bennett, Valts Blukis, Eyvind Niklasson, Max Shatkhin,
  and Yoav Artzi.
\newblock Mapping instructions to actions in 3d environments with visual goal
  prediction.
\newblock In {\em EMNLP}, 2018.

\bibitem{pyrobot2019}
Adithyavairavan Murali, Tao Chen, Kalyan~Vasudev Alwala, Dhiraj Gandhi, Lerrel
  Pinto, Saurabh Gupta, and Abhinav Gupta.
\newblock Pyrobot: An open-source robotics framework for research and
  benchmarking.
\newblock {\em arXiv preprint arXiv:1906.08236}, 2019.

\bibitem{paszke2019pytorch}
Adam Paszke, Sam Gross, Francisco Massa, Adam Lerer, James Bradbury, Gregory
  Chanan, Trevor Killeen, Zeming Lin, Natalia Gimelshein, Luca Antiga, et~al.
\newblock Pytorch: An imperative style, high-performance deep learning library.
\newblock In {\em NeurIPS}, 2019.

\bibitem{pennington_emnlp14}
Jeffrey Pennington, Richard Socher, and Christopher~D. Manning.
\newblock {Glove: global vectors for word representation}.
\newblock In {\em EMNLP}, 2014.

\bibitem{ross2011reduction}
St{\'e}phane Ross, Geoffrey Gordon, and Drew Bagnell.
\newblock A reduction of imitation learning and structured prediction to
  no-regret online learning.
\newblock In {\em AISTATS}, 2011.

\bibitem{schulman2017proximal}
John Schulman, Filip Wolski, Prafulla Dhariwal, Alec Radford, and Oleg Klimov.
\newblock Proximal policy optimization algorithms.
\newblock {\em arXiv preprint arXiv:1707.06347}, 2017.

\bibitem{tan2019learning}
Hao Tan, Licheng Yu, and Mohit Bansal.
\newblock Learning to navigate unseen environments: Back translation with
  environmental dropout.
\newblock In {\em NAACL HLT}, 2019.

\bibitem{vaswani2017attention}
Ashish Vaswani, Noam Shazeer, Niki Parmar, Jakob Uszkoreit, Llion Jones,
  Aidan~N Gomez, {\L}ukasz Kaiser, and Illia Polosukhin.
\newblock Attention is all you need.
\newblock In {\em NeurIPS}, 2017.

\bibitem{wang2019reinforced}
Xin Wang, Qiuyuan Huang, Asli Celikyilmaz, Jianfeng Gao, Dinghan Shen,
  Yuan-Fang Wang, William~Yang Wang, and Lei Zhang.
\newblock Reinforced cross-modal matching and self-supervised imitation
  learning for vision-language navigation.
\newblock In {\em CVPR}, 2019.

\bibitem{wijmans2019dd}
Erik Wijmans, Abhishek Kadian, Ari Morcos, Stefan Lee, Irfan Essa, Devi Parikh,
  Manolis Savva, and Dhruv Batra.
\newblock {DD-PPO}: learning near-perfect pointgoal navigators from 2.5 billion
  frames.
\newblock In {\em ICLR}, 2020.

\bibitem{xia2020relmogen}
Fei Xia, Chengshu Li, Roberto Mart{\'\i}n-Mart{\'\i}n, Or Litany, Alexander
  Toshev, and Silvio Savarese.
\newblock Relmogen: leveraging motion generation in reinforcement learning for
  mobile manipulation.
\newblock {\em arXiv preprint arXiv:2008.07792}, 2020.

\bibitem{yokoyama2021success}
Naoki Yokoyama, Sehoon Ha, and Dhruv Batra.
\newblock Success weighted by completion time: A dynamics-aware evaluation
  criteria for embodied navigation.
\newblock {\em arXiv preprint arXiv:2103.08022}, 2021.

\end{thebibliography}
}

\clearpage
\csection{Supplementary}

\csubsection{Hyperparameter Details}

We use the shared hyperparameters in \tabref{tab:hyperparameters} for all experiments involving RL training. We base our hyperparameter choices on those used by \cite{wijmans2019dd} for PointGoal navigation and adjust specific values as necessary to obtain efficient training convergence on the VLN-CE task. Notably, we find that the slack reward scalar ($r_{slack}$) has a significant impact on training -- changing the default value of \mbox{-0.01} to \mbox{-0.05} leads to reduced reward variance during training, faster convergence, and more efficient paths in both training and validation (a higher \texttt{SPL}). We suspect the vehicle of this change is more cautious multi-step exploration of the state space, which in moderation can be effective in a setting of dense reward with a valid shortest-path-to-goal assumption.

\xhdr{The Discount Factor.}
The action space of the HPN model is effectively a subset of the unconstrained WPN action space. Despite HPN models obtaining higher success metrics than WPN models, WPN models learn different behaviors during training (\eg making distance predictions greater than 0.25m). We find this is a consequence of the discounted reward. In small-scale WPN experiments, setting the discount factor to 1 leads to HPN-style actions whereas setting it to 0 (or values even higher, like 0.5) leads to taking max-distance steps towards the goal. Our experiments use $\gamma=0.99$ to balance immediate gain (emphasizing larger steps) with long-term success.

\begin{table}[t]
	\setlength{\tabcolsep}{3pt}
	\begin{center}
		\begin{tabular}{l c}
			\toprule

			\multicolumn{2}{c}{\textbf{PPO Parameters}} \\
			Parallel simulation environments & 4 \\
			Rollout length (steps per environment) & 16 \\
			DDPPO sync fraction & 0.6\\
			Number of PPO Epochs & 2 \\
			Mini-batches per epoch & 4 \\
			
			Optimizer & Adam \\
			\hspace{0.5cm} Learning rate & 2.0 $\times$ 10$^{-4}$ \\
			\hspace{0.5cm} Epsilon ($\epsilon$) & 1.0 $\times$ 10$^{-5}$ \\
			\hspace{0.5cm} Learning rate decay & False \\

			PPO-clip & 0.2 \\
			\hspace{0.5cm} clip decay & False \\
			Clip the value loss & True \\
			Generalized advantage estimation (GAE) & True \\
			\hspace{0.5cm} Normalized & True \\
			\hspace{0.5cm} $\gamma$ & 0.99 \\
			\hspace{0.5cm} $\tau$ & 0.95 \\
			
            Value loss coefficient ($c_v$) & 0.5 \\
            Offset regularization coefficient ($c_r$) & 0.1146 \\
			Entropy coefficient ($c_e$) & 0.1 \\
            \hspace{0.5cm} \texttt{Pano} entropy coefficient ($c_p$) & 1.5 \\
            \hspace{0.5cm} \texttt{Offset} entropy coefficient ($c_o$) & 1.0 \\
            \hspace{0.5cm} \texttt{Distance} entropy coefficient ($c_d$) & 1.0 \\
			Max gradient norm & 0.2 \\
			
			\midrule

			\multicolumn{2}{c}{\textbf{Reward Parameters}} \\
			Success ($r_{\text{success}}$) & 2.5 \\
			Success distance & 3.0m \\
			slack reward ($r_{slack}$) scalar & -0.05\\

            \bottomrule
		\end{tabular}
	\end{center}
	\caption{Hyperparameters shared by all experiments.}
	\label{tab:hyperparameters}
\end{table}

\begin{figure*}[t]
    \centering
    \includegraphics[width=\textwidth]{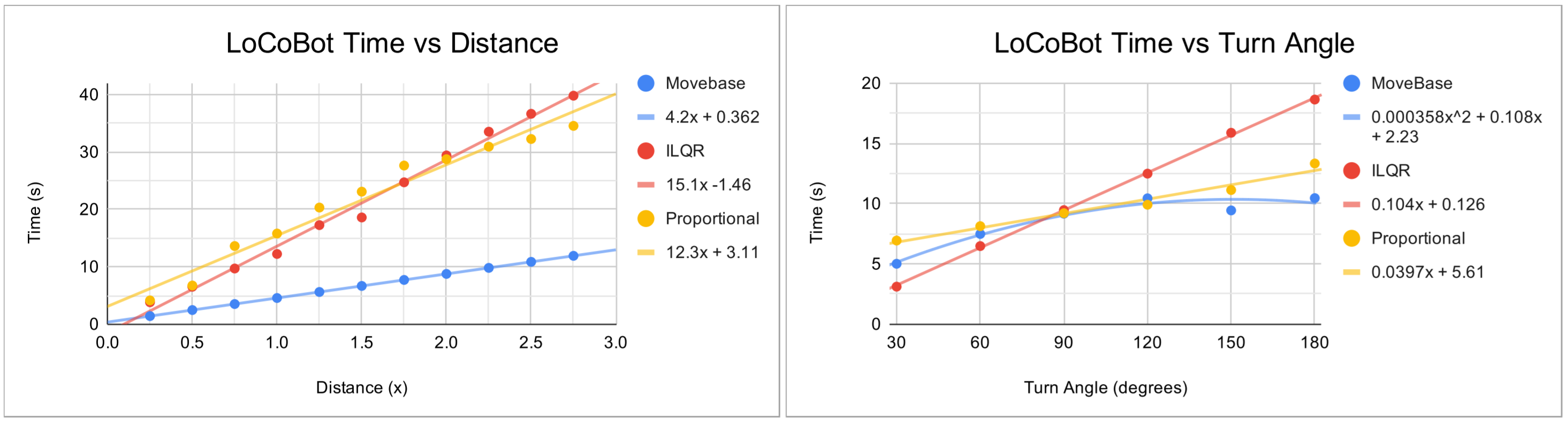}
    \hfill
    \caption{Motion models of a profiled LoCoBot robot. We compute fits for the base controllers MoveBase, ILQR, and Proportional.}
    \label{fig:locobot_fits}
\end{figure*}

\csubsection{Constructing a LoCoBot Motion Model}
In \secref{sec:express}, we estimate the time a LoCoBot robot running our navigation policy would take to execute trajectories in the real world. To support this metric (\texttt{EET}), we determine a motion model from repeated empirical timings of a physical LoCoBot. This model consists of a rotation function that maps turn angle to time and a translation function that maps straight-line distance to time. Together, these functions can estimate execution time for both our continuous navigator and discrete navigator. We profile the base controllers \texttt{MoveBase}, \texttt{ILQR}, and \texttt{Proportional} and show their respective motion models in \figref{fig:locobot_fits}. We find that \texttt{MoveBase} generally leads to faster execution time and use it for all time estimates. We recognize there can be a time \vs accuracy trade-off but for the purposes of \texttt{EET} we are primarily interested in time.

To fit the rotation function, we record the time elapsed between issuing the rotation command and the robot stopping after turning an angle $\phi = 30\degree, 60\degree, ..., 180\degree$. We compute an average time-to-turn by repeating this collection 5 times for each turn angle. We then fit a quadratic function, yielding:
\begin{align}
    y^{\text{rotate}} = 0.000358\phi^2 + 0.108\phi + 2.23.
\end{align}
for \texttt{MoveBase}. We repeat the process for the translation function, collecting 5 timings for each distance $x = 0.25\text{m}, 0.5\text{m}, ..., 2.75\text{m}$. We find a linear fit of:
\begin{align}
    y^{\text{translate}} = 4.2x + 0.362.
\end{align}
Motion models for other robots can be computed in similar fashion.

\csubsection{Adapting Success Weighted by Completion Time (SCT)}
Success weighted by Completion Time (\texttt{SCT}) \cite{yokoyama2021success} is an evaluation metric that scales the agent's episodic binary success by the relative time taken to complete the trajectory. Concretely, \texttt{SCT} is defined as:
\begin{align}
    \texttt{SCT} = \frac{T}{\text{max}(C,T)}
\end{align}
where $C$ is the agent's estimated completion time and $T$ is the minimal time required for an oracle to reach the goal as afforded by the agent's dynamics. While originally designed for unicycle-cart dynamics, we adapt \texttt{SCT} to our experimental settings and report results for each model in \tabref{tab:ablations}. To compute the completion time $C$, we use the estimated execution time (\texttt{EET}) based on our LoCoBot motion model. To determine the minimal time $T$, we adapt the RRT*-Unicycle algorithm to reflect our point-turn dynamics model, our agent's action space, and multi-floor environments. Specifically, we:
\begin{enumerate}
    \item Redefine the cost to travel from one agent pose to another in terms of our LoCoBot motion model,
    \item Sample navigable points up to 4.0m away from existing graph nodes to reflect the action space of our least-constrained WPN model, and
    \item Extend the 2D algorithm to work in multi-floor environments by a) swapping all instances of 2D Euclidean distance with geodesic distance computed on a navigation mesh, and b) sampling, projecting, and interpolating random points from the space of all navigable points.
\end{enumerate}

\begin{figure*}
    \centering
    \resizebox{0.98\textwidth}{!}{
        \begin{tabular}{m{0.125\textwidth} || m{0.17\textwidth} m{0.17\textwidth} m{0.17\textwidth} m{0.17\textwidth} m{0.17\textwidth}}
        \toprule
        TL/EET (cm/s) & \multicolumn{1}{c}{5.0 - 6.5} & \multicolumn{1}{c}{6.5 - 8.0} & \multicolumn{1}{c}{8.0 - 9.5} & \multicolumn{1}{c}{9.5 - 11.0} & \multicolumn{1}{c}{11.0 - 12.5} \\
        \midrule
    
        WPN+CN Episode Maps
         & \includegraphics[width=0.17\textwidth]{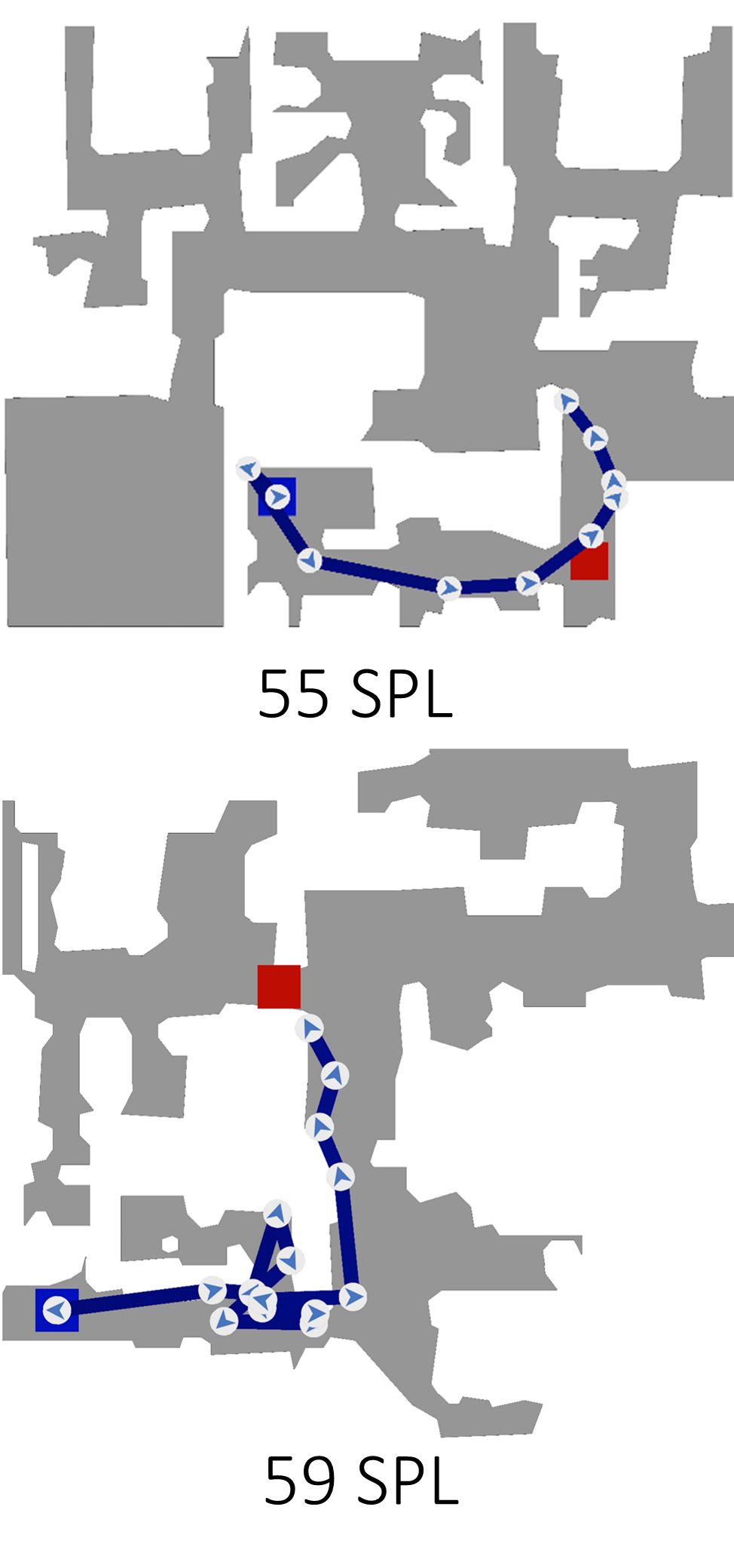}
         & \includegraphics[width=0.17\textwidth]{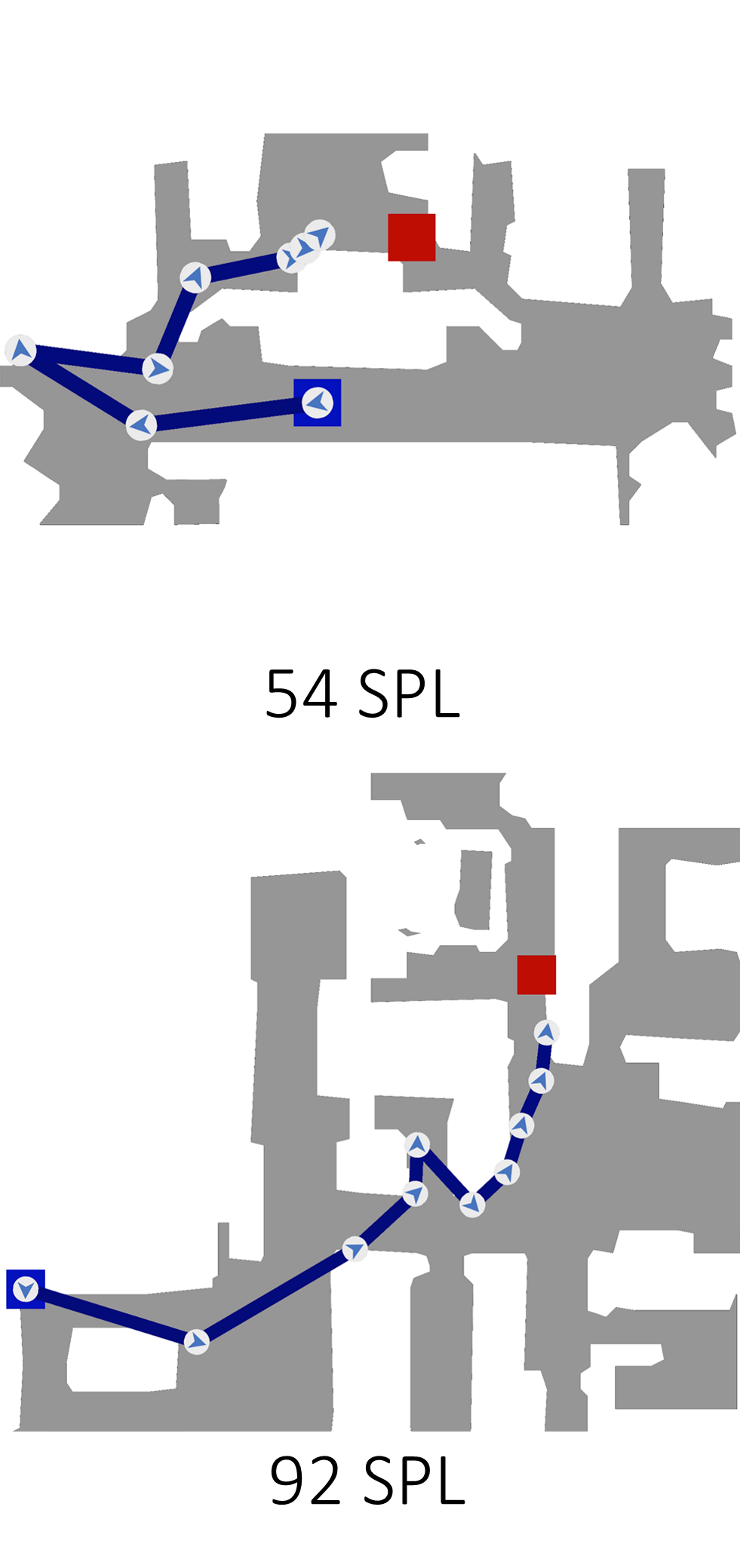}
         & \includegraphics[width=0.17\textwidth]{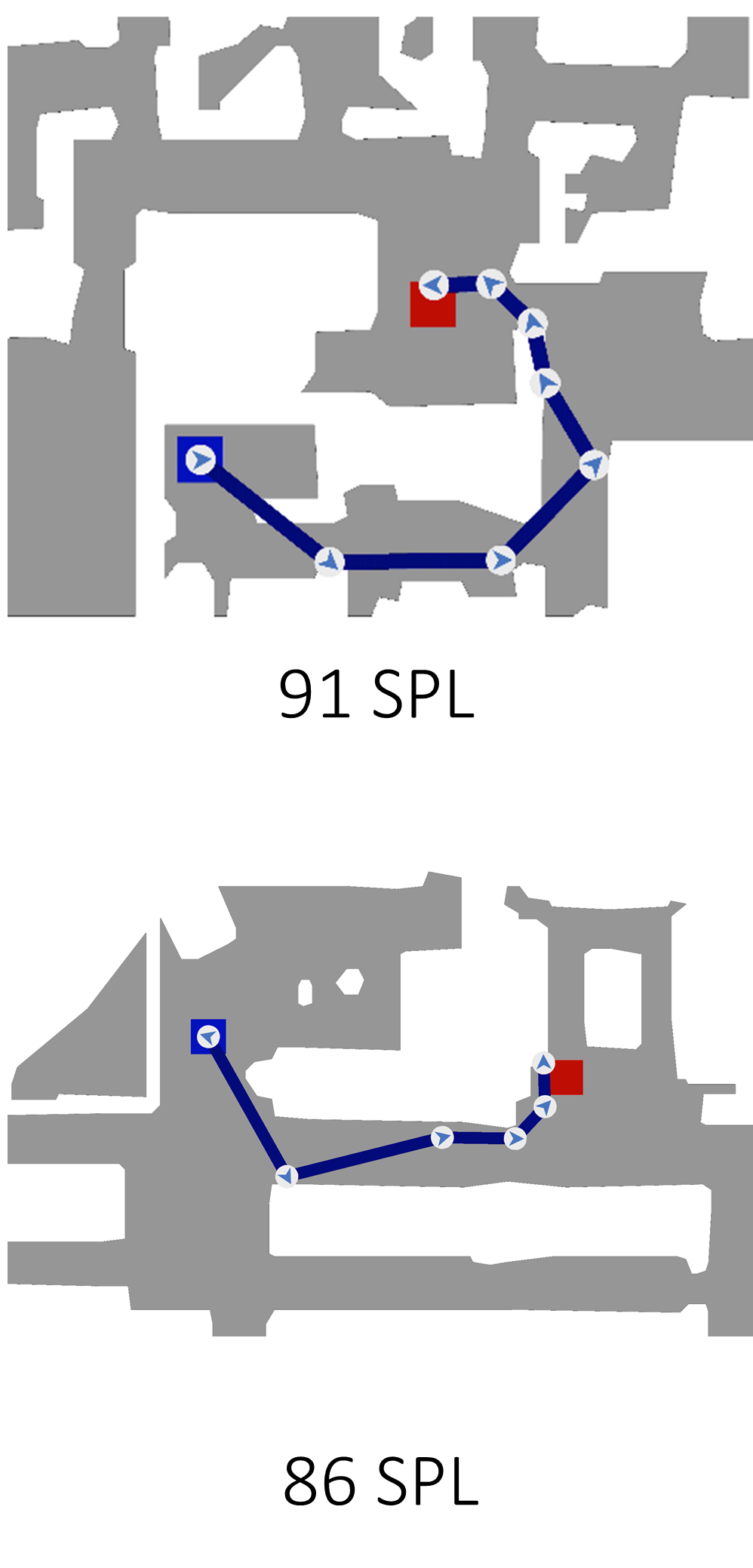}
         & \includegraphics[width=0.17\textwidth]{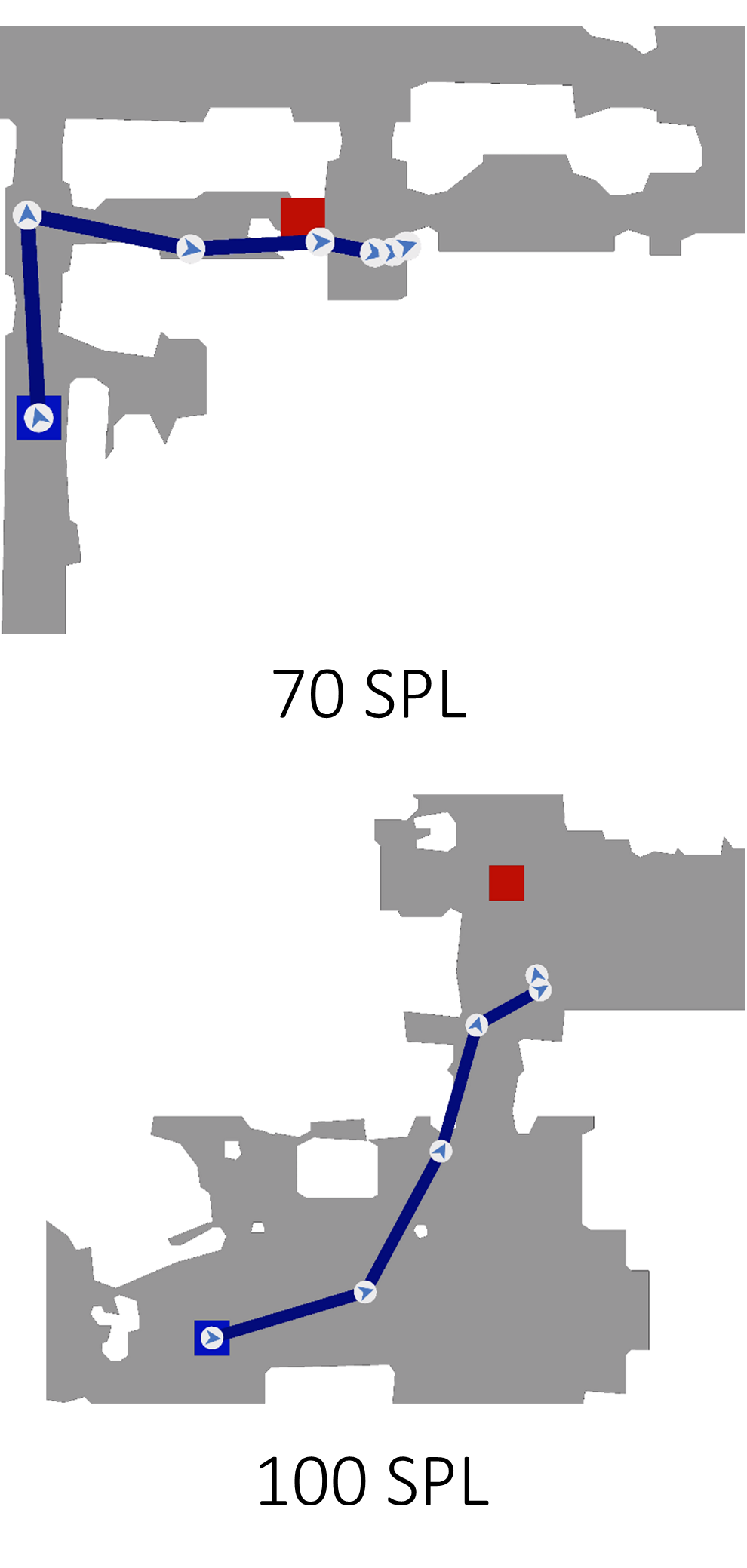}
         & \includegraphics[width=0.17\textwidth]{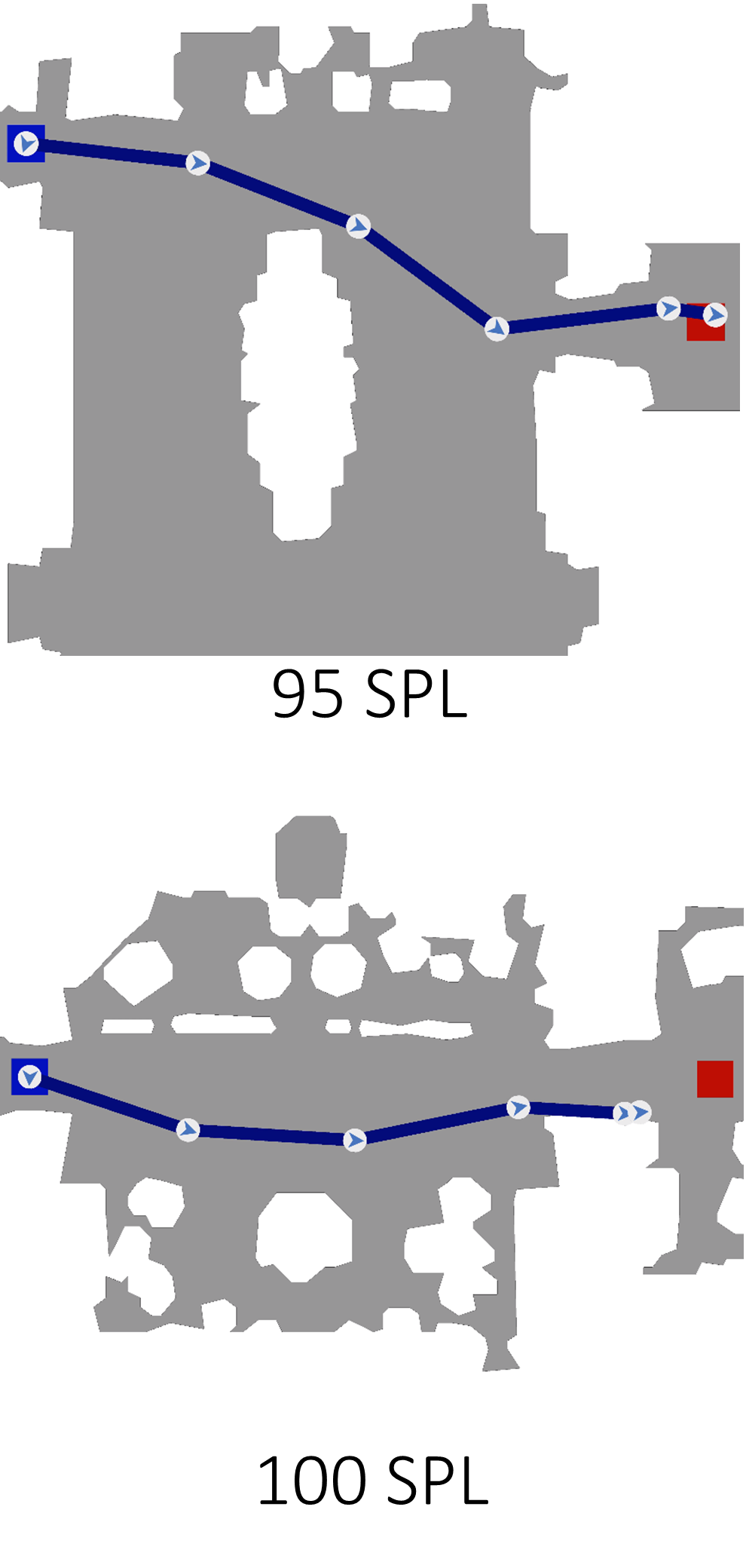}
        \\
        \bottomrule
        \end{tabular}
    }
    \caption{Example episodes of our WPN+CN model in Val-Unseen broken down by the estimated average execution speed (\texttt{TL}/\texttt{EET}). Successful episodes are shown to highlight differences in \texttt{SPL}. Estimated execution time (EET) is calculated from a rotation model and translation model fit to a profiled LoCoBot robot. In each map, the agent starts at the blue square and attempts to navigate to the red square. The agent heading icon denotes the agent's position and heading at the time of each waypoint prediction. Gray regions represent navigable space. These examples show that episodes consisting of short waypoint predictions and large turns lead to slower speed estimates, whereas further waypoints with smaller turns lead to faster speed estimates.}
    \label{fig:eet-examples}
\end{figure*}

\end{document}